\documentclass{article}

\usepackage[preprint]{neurips_2026}

\usepackage[utf8]{inputenc}
\usepackage[T1]{fontenc}
\usepackage{hyperref}
\usepackage{url}
\usepackage{booktabs}
\usepackage{amsfonts}
\usepackage{amsmath,amssymb}
\usepackage{nicefrac}
\usepackage{microtype}
\usepackage[table]{xcolor}
\usepackage{graphicx}
\usepackage{multirow}
\usepackage{array}
\usepackage{enumitem}
\usepackage{subcaption}
\usepackage{pdflscape}
\usepackage{threeparttable}
\usepackage{longtable}
\newtheorem{proposition}{Proposition}[section]

\title{ARBITER: Reasoning Trajectory Basins and Majority Vote Failures in Test-Time Sampling}

\author{%
  Meng Cai \quad Lars Kulik \quad Farhana Choudhury \\
  School of Computing and Information Systems \\
  University of Melbourne \\
  \texttt{meng.cai1@student.unimelb.edu.au} \\
  \texttt{lkulik@unimelb.edu.au} \\
  \texttt{farhana.choudhury@unimelb.edu.au}
}

\begin{document}
\maketitle

\begin{abstract}

When language models use test-time sampling, they generate multiple reasoning trajectories and select an answer by majority vote. We show that these trajectories are not independent: for a given question, they concentrate into a small number of clusters, or \emph{reasoning basins}, each defined by a normalized final answer and the solutions that reach it. A majority vote therefore selects the most stable basin rather than the most accurate one, which creates \emph{wrong-majority failures} where the correct answer is present but outvoted. We introduce \textsc{Arbiter}, a model-agnostic approach that models interactions between basins using only the base model's own sampled outputs, hidden states, and derived evidence. Most direct correction strategies fail; \textsc{Arbiter} instead uses conservative additive evidence on top of consensus.
In its simplest parameter-free form, \textsc{Arbiter-$\Delta$} adds same-model evidence to the majority prior, while \textsc{Arbiter-Enc} augments this with bounded residual signals from hidden states over complete solutions. On GSM8K with Qwen3-4B, consensus over $K{=}24$ samples achieves around the mid-94\% range, while a same-pool top-2 oracle reaches around the mid-96\% range. \textsc{Arbiter} recovers a subset of these cases using zero external information. Across three model families and three math benchmarks, it yields consistent gains with no net-negative cases; for example, on Llama-3.1-8B MMLU-HS-Math, it improves accuracy from the mid-78\% range to the mid-82\% range,
recovering about $22\%$ of the available oracle headroom, indicating that this headroom can be partially recovered from the sample pool itself.
\end{abstract}

\section{Introduction}
\label{sec:introduction}

A standard way to improve language models at inference time is to sample multiple reasoning trajectories and aggregate their final answers by majority vote. This baseline is already strong: across models and benchmarks, it usually improves over greedy decoding and often places the correct answer somewhere in the sampled pool. The remaining challenge is therefore not generation alone, but post-consensus recovery: can a system reliably identify when the majority answer is wrong without degrading the many cases where consensus is already correct?

We find that sampled trajectories concentrate into a small number of coherent answer basins: groups of solutions whose extracted final answers match after task-specific normalization. Majority vote therefore acts as a dominant-basin selector. This usually works well, but an important failure case remains: the dominant basin can be wrong while a correct challenger basin is already present in the sampled pool.

These wrong-majority cases reveal substantial recoverable headroom within the sampled pool itself. In many examples, the correct answer is already present among the observed challenger basins but loses to a larger wrong basin. However, our experiments show that recovering these cases reliably is surprisingly difficult. Broad self-review, hidden-state reranking, trajectory coherence scoring, graph routing, framing-first replacement, and direct basin-selection methods often degrade a strong consensus baseline even when they reveal real structure in the reasoning process. Hidden-state structure and trajectory coherence are therefore not reliable indicators of correctness.

This empirical pattern motivates a more conservative principle: consensus should remain the prior. A challenger basin should override the dominant basin only when additional same-model evidence accumulates in its favor. Reliable recovery comes not from replacing consensus, but from sparse additive evidence layered on top of it.

Existing approaches fall short because they evaluate trajectories in isolation. Sample-and-aggregate methods rely on agreement but do not model relationships between basins. Hidden-state scorers, self-verification methods, and entropy-based approaches assess individual trajectories. They do not directly model comparative evidence between alternative basins. As a result, they struggle to determine when a minority basin should override a stable majority.

We introduce \textit{\textsc{Arbiter}}, a basin-structured framework for post-consensus selection. The key principle is simple: \emph{consensus remains the prior}. It is overridden only when additional same-model evidence supports a challenger basin. For each question, \textsc{Arbiter} groups sampled trajectories into basins and constructs compact representations of their structure. It then collects additional same-model evidence by asking the model to reinterpret, compare, and solve again under competing basin hypotheses, and accumulates this evidence relative to the dominant basin.

Our approach operates in a strict zero-external-information setting. It uses only the model's own sampled outputs and internal representations. It does not rely on external verifiers or additional training signals. This isolates the question of whether the model already contains enough internal evidence to recover from wrong-majority failures.

The central empirical lesson is that post-consensus recovery must be conservative. Many consensus errors are recoverable in principle, but broad correction strategies often harm more correct cases than they fix. Signals such as trajectory coherence and hidden-state structure reveal real organization in the sampled pool, yet they do not reliably identify correctness. Reliable gains instead come from sparse, high-precision overrides supported by additive basin-level evidence.

\medskip
\noindent Our contributions are as follows.

\begin{enumerate}[leftmargin=*, label=\textbf{(\arabic*)}]

\item We identify \emph{wrong-majority failure} as a key failure case of consensus decoding. Sampled trajectories concentrate into a small number of reasoning basins, and majority vote selects the most stable basin rather than the most accurate one.

\item We introduce \textit{\textsc{Arbiter}}, a basin-structured framework that performs post-consensus selection by accumulating same-model evidence for challenger basins relative to the dominant basin, while treating consensus as a prior.

\item We show that post-consensus recovery is inherently selective. Across a broad range of self-review, hidden-state, graph-routing, and framing-based interventions, most direct correction strategies degrade a strong consensus baseline. Reliable recovery instead comes from sparse, high-precision additive evidence.

\end{enumerate}

\section{Related work}
\label{sec:related}

\noindent\textbf{Test-time sampling and answer aggregation.}
Chain-of-thought prompting and self-consistency established the standard recipe of sampling multiple reasoning traces and aggregating final answers~\citep{wei2022cot,wang2022selfconsistency}. Universal Self-Consistency extends this idea by using the model itself to select among candidate solutions beyond exact-answer majority voting~\citep{chen2023usc}. More recent test-time scaling work studies how to allocate inference compute across problems rather than spend it uniformly~\citep{snell2024scaling}. These methods motivate our baseline: raw consensus is a strong dominant-basin estimator. \textsc{Arbiter} keeps that estimator as the prior and studies when same-model evidence justifies selecting another observed basin.

\noindent\textbf{Hidden-state and trajectory signals.}
Prior work uses hidden states, token uncertainty, step-level pruning, latent actions, or proactive refinement to evaluate or control reasoning~\citep{liang2026step,ghasemabadi2024gnosis,chen2026instability,entropy2025branching,han2025stitch,shi2026stir}. Recent work also studies semantic or latent structure across sampled reasoning trajectories, including semantic consistency, latent majority-set selection, and hidden-state clustering approaches~\citep{semanticselfconsistency2024,latentselfconsistency2025,clue2025}. Recent trajectory-level views further support treating a complete solution as a path through latent computation rather than as a bag of isolated token states~\citep{liang2026step,shi2026stir}. This literature supports the idea that model-internal computation contains useful structure. Our results sharpen an important limitation: structure is not truth. Coherence, stability, and graph reconstruction often detect commitment or risk, not correctness. We therefore treat trajectory encoders and basin graphs as residual or diagnostic components, not as standalone selectors.

\noindent\textbf{Self-correction limitations.}
Iterative self-feedback and reflection frameworks show that model-generated feedback can improve outputs in some settings~\citep{madaan2023selfrefine,shinn2023reflexion}. A parallel line of work shows that unguided self-correction can be weak or harmful without a reliable verifier~\citep{huang2024large,zhang2024smalllm,vasudev2026failure}. This matches our broader experimental findings: broad self-review, cluster judging, and direct replacement policies often disrupt already-correct consensus answers. \textsc{Arbiter} responds by making correction sparse, additive, and auditable through recovered/degraded counts.

\noindent\textbf{Framing and semantic decomposition.}
Math-reasoning benchmarks and perturbation studies show that wording, entities, units, and symbolic form can strongly affect model behavior~\citep{cobbe2021gsm8k,hendrycks2021mmlu,hendrycks2021math,li2024gsmplus,mirzadeh2024gsmsymbolic}. We use same-model semantic descriptions to expose competing interpretations of observed answer basins. Unlike framing-first replacement, \textsc{Arbiter-$\Delta$} uses these descriptions only to collect additive evidence on top of the raw consensus prior.

\section{Problem setup}
\label{sec:problem_setup}

We study \emph{post-consensus recovery} for a frozen autoregressive language model $M$ under \emph{zero external information}: the selector uses only the model's own sampled outputs, internal states, and evidence derived from them. Gold labels are used only after prediction for evaluation. For each question $q$, the raw baseline is ordinary sampled generation, final-answer clustering, and majority selection.

The following notation defines the objects used throughout the paper. A \emph{candidate solution} is one complete generated completion. A \emph{trajectory} is used only when hidden states are involved: it is the layer-by-layer, token-by-token hidden-state sequence recorded while the frozen model generates that complete solution. An \emph{answer basin} is an observed final-answer cluster together with the generated solutions and their hidden-state trajectories. We define basins by final-answer agreement, not by requiring all solutions in a basin to share the same reasoning path. The dominant basin is the largest answer basin; challenger basins are all other observed basins. Eqs.~(D1)--(D10) give the formal objects. For reference, Appendix~\ref{app:symbol_reference} lists every symbol used in these definitions and in the method score.

\begin{flalign}
& \mathcal{S}(q) = \{s_1,\ldots,s_K\},\quad s_i=(y_{i,1},\ldots,y_{i,T_i})
&& \text{raw candidate pool} \tag{D1}\\
& H_i = \bigl(h_{i,1}^{(1:L)},\ldots,h_{i,T_i}^{(1:L)}\bigr),\quad
h_{i,t}^{(\ell)}\in\mathbb{R}^{d_{\mathrm{model}}}
&& \text{hidden-state trajectory} \tag{D2}\\
& a_i = \mathrm{Ans}(s_i)
&& \text{task-normalized final answer} \tag{D3}\\
& C_r(q) = \{\,i:a_i=\alpha_r\,\},\quad |C_1|\ge |C_2|\ge\cdots\ge |C_{m(q)}|
&& \text{ranked answer clusters} \tag{D4}\\
& B_r(q) = \bigl(\alpha_r, C_r, \{H_i:i\in C_r\}\bigr)
&& \text{observed answer basin} \tag{D5}\\
& \hat y_{\mathrm{cons}}(q) = \alpha_1
&& \text{raw consensus prediction} \tag{D6}
\end{flalign}

Equations~(D1)--(D6) separate the \emph{answer} selected by consensus from the \emph{solutions} that produced it. Ties in Eq.~(D4) are broken deterministically by the earliest sampled index in the cluster and then by the canonical answer string. A question belongs to the disagreement slice when $m(q)\ge2$. Only such questions admit within-pool arbitration, because no alternative observed basin exists when all sampled solutions collapse to one answer.

Let $y^\star(q)$ be the gold answer. Gold is unavailable to the selector. It is used only to compute accuracy, diagnostic oracle ceilings, and recovered/degraded counts:

\begin{flalign}
& \mathrm{Acc}_{\mathrm{cons}} =
\mathbb{E}_{q}\bigl[\mathbf{1}\{\alpha_1(q)=y^\star(q)\}\bigr]
&& \text{raw consensus accuracy} \tag{D7}\\
& \mathrm{Oracle@}k(q) =
\mathbf{1}\{\exists r\le k:\alpha_r(q)=y^\star(q)\}
&& \text{diagnostic same-pool ceiling} \tag{D8}\\
& \mathrm{WM}(q) =
\mathbf{1}\{\alpha_1(q)\ne y^\star(q),\;\exists r>1:\alpha_r(q)=y^\star(q)\}
&& \text{wrong-majority indicator} \tag{D9}\\
& \Delta\mathrm{Acc}(\pi) =
\mathbb{E}_{q}\!\left[
\mathbf{1}\{\alpha_{\pi(q)}(q)=y^\star(q)\}
-
\mathbf{1}\{\alpha_1(q)=y^\star(q)\}
\right]
&& \text{net recovery of policy $\pi$} \tag{D10}
\end{flalign}

Oracle@$k$ is not a deployable method: it is a diagnostic ceiling that shows how often the correct answer already appears among the observed challenger basins. Equivalently, Eq.~(D10) is the probability of wrong-to-right recoveries minus the probability of right-to-wrong degradations, which explains why high-accuracy consensus is hard to improve. Thus a useful policy must treat consensus as the default and override it only when sufficient same-model evidence accumulates in favor of a challenger basin.

\section{Method}
\label{sec:method}

\subsection{Inference pipeline}
\label{sec:method_pipeline}

\textsc{Arbiter} is a post-consensus arbitration method. For each question, it first samples a raw pool of ordinary solutions, clusters them by final answer, and treats the largest basin as the consensus prior. It then asks the same frozen model to produce compact interpretations of competing basins and to generate auxiliary evidence streams. Each auxiliary output is parsed back into one of the observed answer basins. Finally, \textsc{Arbiter-$\Delta$} adds the evidence as log ratios and keeps consensus unless a challenger has positive accumulated evidence over the dominant basin.

This pipeline is intentionally simple: \emph{sample, cluster, describe, collect evidence, add evidence}. It uses no external verifier, no tool, no human feedback, and no gold label during selection.

\subsection{Evidence sources and notation}
\label{sec:evidence_sources}

A \emph{frame} is a one-sentence semantic interpretation of a basin: what quantity is being asked for, which entities and units are involved, and what operation pattern connects them. A \emph{framed-pool solve} asks the model to state such an interpretation before solving. A \emph{panel trial} shows the model two basin interpretations side by side and then asks it to solve fresh. A \emph{guided re-solve} gives one basin interpretation as a hypothesis and asks the model to re-derive the answer.

For a challenger basin $B_r$ and the dominant basin $B_1$, every evidence source produces counts for the two basins. A count increases when the parsed final answer from that source matches the basin answer. Invalid outputs or answers outside the compared pair are not added to either count, but they reduce the source reliability term below.

For reference, Appendix~\ref{app:symbol_reference} gives the count-level symbol table for these sources.

The rule has no learned or tuned source scaling constants. The smoothing value $\alpha=1$ is a fixed Laplace count used only to avoid zero-count log ratios; it is not a learned prior or a source-weight parameter. For the main top-2 policy, framed and guided reliability are the top-2 masses:
\begin{flalign}
&& r_f(q)=\frac{f_1+f_2}{N_F^{\mathrm{att}}},
\qquad
r_g(q)=\frac{g_1+g_2}{N_G^{\mathrm{att}}},
&& \tag{M1}
\end{flalign}
where $N_F^{\mathrm{att}}$ and $N_G^{\mathrm{att}}$ count all attempted framed and guided outputs, including invalid outputs and answers outside the compared pair. The mass shrinkage means that if many framed or guided trials land on answers other than $B_1$ or $B_2$, that source is unreliable for this question and contributes less. Panel evidence is not part of the main framed+guided rule, but is retained for source-set ablations. For those ablations, let $p_j^+$ and $p_j^-$ be counts for basin $B_j$ when the two frames are shown in the original and swapped order, so $p_j=p_j^++p_j^-$. Let $N_P^{+,\mathrm{att}}$ and $N_P^{-,\mathrm{att}}$ be the attempted counts in the two orders. We use top-pair mass times order symmetry:
\begin{flalign}
&& m_{P,r}=\frac{p_1^+ + p_r^+ + p_1^- + p_r^-}{N_P^{+,\mathrm{att}}+N_P^{-,\mathrm{att}}},
\quad
\nu_r^{\pm}=\frac{p_r^{\pm}+\alpha}{p_1^{\pm}+p_r^{\pm}+2\alpha},
\quad
\rho_{P,r}=m_{P,r}\bigl(1-|\nu_r^+-\nu_r^-|\bigr).
&& \tag{M1p}
\end{flalign}
Thus $\rho_{P,r}$ is small when panel outputs go off-pair or when swapping the order of the two shown frames changes the top-pair distribution substantially. These factors use only model outputs and parse statistics, never gold labels.

\subsection{The Delta rule}
\label{sec:arbiter_delta}

For each question $q$ with at least two observed basins, let $B_1$ be the majority basin and $B_2$ the leading challenger. The main parameter-free \textsc{Arbiter-$\Delta$} policy uses only raw, framed, and guided counts for this top-2 pair:
\begin{flalign}
& \small \Delta_2(q)
= \log\frac{b_2+\alpha}{b_1+\alpha}
+ r_f(q)\log\frac{f_2+\alpha}{f_1+\alpha}
+ r_g(q)\log\frac{g_2+\alpha}{g_1+\alpha}.
&& \tag{M2}
\end{flalign}
The first term is the raw majority prior: the challenger starts behind when it has fewer raw votes. The second and third terms are same-model framed-pool and guided re-solve evidence. The fixed pseudo-count is $\alpha=1.0$. There is no panel term in the main policy; panel evidence is used only in source-set ablations. The rule selects $B_2$ iff $\Delta_2(q)>0$ and otherwise keeps $B_1$:
\begin{flalign}
& \hat y(q)=
\begin{cases}
\alpha_2, & \Delta_2(q)>0,\\[2pt]
\alpha_1, & \text{otherwise.}
\end{cases}
&& \tag{M3}
\end{flalign}
For notational compatibility with appendix ablations, we write $\Delta_r$ for the analogous score comparing a challenger $B_r$ against $B_1$; the main reported policy uses $r=2$. Thus the decision boundary is simply the sign of the accumulated evidence. The framed+guided source set was fixed before the $3\times3$ evaluation matrix was reported; panel and all-source variants are retained only as source-set ablations. The same sign rule and the same fixed main source set are applied across the evaluation matrix. In practice, almost all accepted moves are to the leading challenger: weaker basins rarely accumulate enough evidence to overcome the raw-consensus prior, and their oracle contribution is usually small. In the Qwen3-4B GSM8K run, raw consensus is $94.54\%$, the top-2 oracle is $97.27\%$, the top-3 oracle is $97.88\%$, and top-5 adds only limited additional headroom. Appendix~\ref{app:worked_delta} gives a complete count-level calculation of Eq.~(M2).

\subsection{Log-linear source-pooling view}
\label{sec:bayes_view}

Equation~(M2) is a Bayes-motivated log-linear pooling score, not a calibrated Bayesian posterior over all auxiliary samples. For the top-2 pair, define unnormalized pooled support for $j\in\{1,2\}$ as
\[
\widetilde w_j(q)
=(b_j+\alpha)\bigl(f_j+\alpha\bigr)^{r_f(q)}\bigl(g_j+\alpha\bigr)^{r_g(q)}.
\]
Taking the log ratio gives the implemented score exactly:
\begin{flalign}
&& \log\frac{\widetilde w_2(q)}{\widetilde w_1(q)}
= \log\frac{b_2+\alpha}{b_1+\alpha}
+ r_f(q)\log\frac{f_2+\alpha}{f_1+\alpha}
+ r_g(q)\log\frac{g_2+\alpha}{g_1+\alpha}
=\Delta_2(q).
&& \tag{M4}
\end{flalign}
Thus Eq.~(M3) selects the challenger exactly when its pooled support exceeds the dominant basin's pooled support. The form keeps the Bayesian intuition that log evidence adds, but it deliberately treats each evidence source as one bounded empirical opinion. A full Bayesian update would generally scale with the number of auxiliary trials; Eq.~(M2) avoids that scaling because same-model evidence streams can be correlated. Appendix~\ref{app:delta_derivation} gives the exact algebraic derivation of this pooling identity.

\begin{figure}[!h]
\centering
\IfFileExists{fig_basin_combined.pdf}{%
  \includegraphics[width=\linewidth]{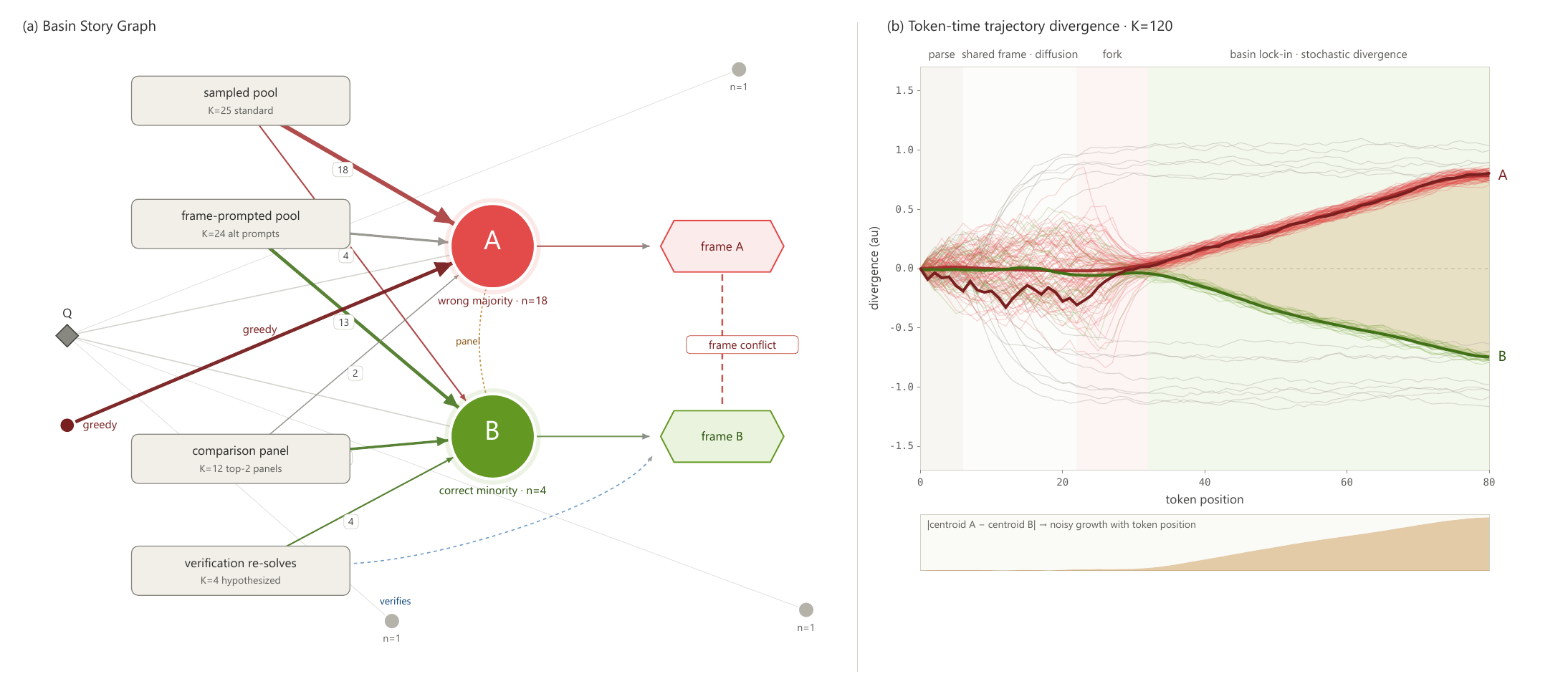}%
}{%
  \fbox{\parbox{0.92\linewidth}{\centering Placeholder for \texttt{fig\_basin\_combined.pdf}. Upload the figure file in Overleaf to render this panel.}}%
}
\caption{One wrong-majority case for \textsc{Arbiter-$\Delta$}.
\textbf{(a)} Basin Story Graph: dominant basin $B_1{=}A$ ($n{=}18$), challenger $B_2{=}B$ ($n{=}4$), three singleton basins. Edges from each evidence source (\S\ref{sec:evidence_sources}) to the top-2 are weighted by the per-basin counts $(b,f,p,g)$; the dashed red edge marks the conflicting frames between $B_1$ and $B_2$. The sampled pool favors $A$ ($b_1{=}18,b_r{=}4$), but the compact framed+guided main policy already favors $B$ ($f{:}13{\,\text{vs.}\,}8$, $g{:}4{\,\text{vs.}\,}0$), and the panel stream gives additional support in full-source ablations ($p{:}9{\,\text{vs.}\,}2$); $\Delta_r{>}0$ in Eq.~(M2), so the rule flips selection to $B$.
\textbf{(b)} Hidden-state divergence visualization for the same example; the larger $K{=}120$ visualization is for illustrative only, not a quantitative reference.
}
\label{fig:basin_combined}
\end{figure}

\subsection{Optional encoder residual}
\label{sec:arbiter_enc}

\textsc{Arbiter-Enc} is a diagnostic upper-bound variant, not the main claim: it indicates how much additional headroom is reachable when a bounded hidden-state residual is added on top of Eq.~(M2). It is not trained to say which answer is correct. Its objectives are: predict held-out same-model evidence, estimate whether the residual is reliable enough to use, and keep the correction small enough that it cannot overwhelm the hand Delta rule.

For basin $B_r$, the encoder aggregates the hidden-state trajectories of the solutions in that basin into a representation $z_r=A_\theta(\{H_i:i\in C_r\})$, where $A_\theta$ is a permutation-invariant trajectory aggregator. It outputs a signed residual $e_{\theta,r}(q)$ and, when enabled, a permission score $\gamma_{\theta,r}(q)\in[0,1]$:
\begin{flalign}
&& \Delta^{\mathrm{Enc}}_r(q)
=
\Delta_r(q)
+
\lambda\,\gamma_{\theta,r}(q)\,\operatorname{clip}\!\left(e_{\theta,r}(q),-c_{\mathrm{clip}},c_{\mathrm{clip}}\right).
&& \tag{M5}
\end{flalign}
The clipped form makes the encoder residual conservative by construction. Here $\lambda\ge0$ is the residual scale and $c_{\mathrm{clip}}>0$ is the clipping bound; both are fixed before evaluation in the fixed setting and may be tuned only on labeled validation data in the calibrated setting. Runs that do not use a permission head set $\gamma_{\theta,r}(q)\equiv1$. We report calibrated residuals separately because they are a stronger but different regime.

Training uses held-out evidence rather than gold labels. For example, the encoder may see raw and framed-pool features and learn to predict the panel/guided evidence delta. The basic target is:
\begin{flalign}
&& \delta_{o,r}(q)=\log\frac{n_{o,r}+\alpha}{n_{o,1}+\alpha},
\qquad
\mathcal{L}_{\mathrm{res}}=\sum_{q,r>1}\omega_{q,r}\,\mathrm{Huber}\!
\left(e_{\theta,r}(q)-\delta_{o,r}(q)\right),
&& \tag{M6}
\end{flalign}
where $o$ is the held-out evidence source or source group, $n_{o,j}$ is the number of held-out outputs from $o$ assigned to basin $B_j$, and $\omega_{q,r}\in[0,1]$ is a prespecified weight, typically the held-out top-pair mass $(n_{o,1}+n_{o,r})/N_o^{\mathrm{att}}$. We use the standard Huber loss, which is quadratic for small residuals and linear for large residuals, so that a small number of noisy held-out counts cannot dominate training. If a permission head is learned, it is trained only from held-out evidence reliability and is used to attenuate this residual, not to predict gold correctness. In words, the encoder learns whether independent same-model evidence would support a challenger; it does not learn a direct correctness label.

\section{Experiments}
\label{sec:experiments}

We evaluate post-consensus recovery under zero external information. The
section is organized around the empirical path of the project: establish raw
consensus and oracle headroom, summarize the negative correction attempts that
motivated the framing pivot, and report the main positive results for
\textsc{Arbiter-$\Delta$} and \textsc{Arbiter-Enc}. Full per-experiment
details, historical runs, and reliability tiers are in
Appendices~\ref{app:full_ledger}--\ref{app:artifact_map}.

\textbf{Methods:} The baseline throughout is \emph{raw consensus}: a $K{=}24$ ordinary sampled
solution pool, answer clustering, and majority selection. Reported $K$ counts the sampled generations used in the consensus vote; a greedy anchor is recorded separately for greedy accuracy and compute accounting and is not included in the raw-consensus vote. We use $K{=}24$ as the default raw pool size. A GSM8K raw-only diagnostic at $K{=}56$ did not outperform $K{=}24$ for the three model families (Appendix~\ref{app:full_ledger}, Table~\ref{tab:gsm8k_k56_raw}), but we do not treat it as a strict compute-matched full-matrix baseline. This baseline uses no framing
prompt, no frame panel, no guided re-solving, and no encoder-based selection.
Because different scripts use different seeds, prompt families, and run
configurations, every method is compared only to the raw consensus baseline
from the same run family. The methods are summarized in Table~\ref{tab:evidence_budget}.

\textbf{Metrics:} For every selector, we report accuracy and the correction decomposition: \emph{overrides}, \emph{recovered}, \emph{degraded}, and \emph{net = recovered $-$ degraded}. Recovered cases are wrong raw-consensus predictions changed to correct; degraded cases are correct raw-consensus predictions changed to wrong. This decomposition is essential because raw consensus is already strong, and a selector can appear plausible while hurting more correct cases than it fixes. At high consensus accuracies, even modest positive gains are difficult to obtain without introducing larger numbers of right-to-wrong degradations. The benchmark size $N$ is not reduced by model parse failures; invalid generated outputs are treated as invalid predictions/evidence and remain in attempted-output denominators. We also report the evidence budget: raw-pool size, number of framed solves, number of panel trials, number of guided re-solves, and whether an encoder residual is used. Across all tested seeds and run families, \textsc{Arbiter}-$\Delta$ exceeds raw consensus on this decomposition.

\begin{table}[t]
\centering
\caption{Evidence budget and status of the main methods. Framed-pool solves, panel trials, and guided trials are defined in Section~\ref{sec:evidence_sources}; all auxiliary evidence is generated by the same frozen base model.}
\label{tab:evidence_budget}
\footnotesize
\setlength{\tabcolsep}{4pt}
\begin{tabular}{p{2.6cm}p{4.8cm}p{4.1cm}}
\toprule
\textbf{Method} & \textbf{Evidence budget} & \textbf{Status} \\
\midrule
Raw consensus & $K{=}24$ ordinary sampled solutions; answer clustering; majority selection. & Clean zero external information. \\
Framed-pool consensus & Additional framed generations from the same base model. & Ablation; not the raw baseline. \\
\textsc{Arbiter-$\Delta$} & Raw pool + fixed framed-pool and frame-guided evidence; no panel term in the main Delta score. Panel evidence is reported as a source-set ablation. & Main clean zero-external-information method. \\
\textsc{Arbiter-Enc-Fixed} & \textsc{Arbiter-$\Delta$} + fixed bounded trajectory residual. & Clean zero external information when the residual rule is fixed before evaluation. \\
\textsc{Arbiter-Enc-Calibrated} & \textsc{Arbiter-$\Delta$} + validation-calibrated encoder residual. & Calibrated; not strict zero external information. \\
\bottomrule
\end{tabular}
\end{table}

\textbf{Datasets and language models:} We evaluate frozen instruction-tuned language models across math reasoning
benchmarks. The main multi-model results use (i) Qwen3-4B-Instruct-2507~\citep{yang2025qwen3}, (ii)
 Llama-3.1-8B-Instruct~\citep{dubey2024llama3}, and (iii) Phi-4~\citep{abdin2024phi4}
over the datasets of (i) GSM8K~\citep{cobbe2021gsm8k} ($n{=}1319$), (ii) MMLU-HS-Math~\citep{hendrycks2021mmlu}
($n{=}270$), and (iii) MATH-500~\citep{hendrycks2021math,lightman2023letsverify}
($n{=}500$). GSM8K answers are normalized numeric strings, MMLU-HS-Math answers are
canonical multiple-choice labels, and MATH-500 answers are normalized boxed expressions
with symbolic equivalence when available.

\subsection{Experiment results}
\textbf{Raw consensus and oracle headroom:} 
The full consensus/oracle table is in Appendix~\ref{app:full_ledger} (Table~\ref{tab:app_consensus_oracle}). The results show that raw consensus leaves visible same-pool headroom. In brief, consensus improves over greedy in seven of nine model--dataset cells, matches it in one cell, and remains the in-family baseline for all selector comparisons; the single below-greedy cell is reported explicitly in the appendix. In that Qwen3-4B MATH-500 cell, the same-pool oracle remains higher than both greedy and consensus, indicating that the 4B model often samples a correct answer but does not form a stable dominant correct basin. The top-3 oracle reveals substantial recoverable headroom, especially for the models that have lower accuracy and harder benchmarks. A GSM8K raw-only diagnostic at $K{=}56$ did not improve over $K{=}24$ (Table~\ref{tab:gsm8k_k56_raw}), so the gains below should not be interpreted as merely drawing more ordinary samples.

\textbf{Negative ladder: most global correction fails:} 
The full negative ladder is reported in Appendix~\ref{app:full_ledger}. Broad self-review, top-up vote merging, raw-trace review, answer-memo review, basin-principle judging, our original trajectory encoder, the cluster-GNN/router, framing-first replacement, and direct panel/guided replacement all fail as global selectors. These results are not incidental: they show that stability, coherence, and graph structure often measure self-consistency rather than correctness. This motivates the final design choice: keep raw consensus as the prior and add auxiliary evidence only through a conservative arbitration score.

\textbf{Main result: \textsc{Arbiter-$\Delta$}:} 
\label{sec:exp_delta_main}
Table~\ref{tab:delta_main} reports the main clean result using the fixed framed+guided parameter-free policy. Each row compares
\textsc{Arbiter-$\Delta$} against the raw consensus baseline from the same
run family. Count-level net change is positive in 8 of 9 cells and neutral
in one; accuracy is positive in eight cells and unchanged in one. Net counts are recovered minus degraded examples.

\begin{table}[htbp]
\centering
\caption{\textsc{Arbiter-$\Delta$} across three models and three benchmarks using the fixed framed+guided parameter-free policy. Raw consensus is the in-family no-framing majority baseline. ``Overrides / Rec. / Deg.'' means arbitration moves, wrong-to-right recoveries, and right-to-wrong degradations. Counts are reported for every benchmark row.}
\label{tab:delta_main}
\footnotesize
\setlength{\tabcolsep}{4pt}
\begin{tabular}{llrrrrlr}
\toprule
\textbf{Model} & \textbf{Dataset} & \textbf{N}
& \textbf{Raw cons.} & \textbf{\textsc{Arbiter-$\Delta$}}
& \textbf{Gain (pp)} & \textbf{Overrides / Rec. / Deg.} & \textbf{Net} \\
\midrule
Llama-3.1-8B & GSM8K        & 1319 & 91.81 & 92.65 & +0.84 & 35 / 20 / 9 & +11 \\
Llama-3.1-8B & MMLU-HS-Math & 270  & 78.52 & 80.00 & +1.48 & 17 / 9 / 5 & +4 \\
Llama-3.1-8B & MATH-500     & 500  & 51.60 & 54.60 & +3.00 & 44 / 19 / 4 & +15 \\
\rowcolor{gray!10}
\multicolumn{3}{l}{\textit{Llama-3.1-8B mean}} & 73.98 & 75.75 & \textbf{+1.77} & & \textbf{+10.0} \\
\midrule
Qwen3-4B & GSM8K        & 1319 & 94.54 & 94.84 & +0.30 & 16 / 9 / 5 & +4 \\
Qwen3-4B & MMLU-HS-Math & 270  & 94.81 & 95.19 & +0.37 & 10 / 3 / 2 & +1 \\
Qwen3-4B & MATH-500     & 500  & 69.20 & 69.20 & +0.00 & 11 / 4 / 4 & 0 \\
\rowcolor{gray!10}
\multicolumn{3}{l}{\textit{Qwen3-4B mean}} & 86.18 & 86.41 & \textbf{+0.23} & & \textbf{+1.7} \\
\midrule
Phi-4 & GSM8K        & 1319 & 96.13 & 96.44 & +0.31 & 9 / 6 / 2 & +4 \\
Phi-4 & MMLU-HS-Math & 270  & 92.59 & 93.70 & +1.11 & 4 / 3 / 0 & +3 \\
Phi-4 & MATH-500     & 500  & 73.00 & 73.20 & +0.20 & 22 / 5 / 4 & +1 \\
\rowcolor{gray!10}
\multicolumn{3}{l}{\textit{Phi-4 mean}} & 87.24 & 87.78 & \textbf{+0.54} & & \textbf{+2.7} \\
\bottomrule
\end{tabular}
\end{table}

The largest gains occur where consensus has more headroom: Llama-3.1-8B averages $+1.77$ points across datasets, with the strongest single-cell gain of $+3.00$ on MATH-500. Near-ceiling cells show small or neutral changes, consistent with the oracle analysis. The Qwen3-4B GSM8K policy illustrates the operating regime: 16 overrides yield 9 recoveries and 5 degradations. Across the evaluation matrix, the method makes 168 overrides, 78 recoveries, and 35 degradations, for a net gain of 43 examples. The method improves by making a small number of higher-precision non-consensus moves, not by broadly rewriting predictions.

\textbf{\textsc{Arbiter-Enc}: encoder residual gains:} 
The trajectory encoder is a bounded residual on top of \textsc{Arbiter-$\Delta$}; it is not a standalone verifier and is not expected to improve the fixed raw-majority vote itself. Appendix~\ref{app:enc_ranges} reports the full fixed/no-tune and calibrated/tuned gain ranges. The fixed encoder is near-zero to modest in the strict label-free setting, mainly by reducing degraded flips and identifying noisy framing evidence. The calibrated encoder shows a stronger range when labeled validation data is available, so we report it separately from the strict zero-external-information setting.

\textbf{Experimental results overview:}
The full empirical-program table is in Appendix~\ref{app:program_overview} (Table~\ref{tab:experiment_program}). The overview of the results is: raw consensus is the stable sampled-vote baseline, most global correction mechanisms are negative or fragile, and the clean fixed full-matrix selector with non-negative count-level net outcomes is \textsc{Arbiter-$\Delta$}. Unless stated otherwise, representative GSM8K numbers in the appendix use Qwen3-4B on the GSM8K test split.

\textbf{Summary:} 
Five findings follow from the experiments.
(1) Raw consensus is the correct in-family baseline -- it improves or matches greedy in most evaluated cells and exposes substantial recoverable headroom.
(2) Most global correction methods fail or are fragile.
(3) Framing helps only as additive evidence; framing-as-replacement is
negative.
(4) \textsc{Arbiter-$\Delta$} has non-negative point-estimate count-level net change in all 9 cells and positive accuracy gain in eight cells.
(5) \textsc{Arbiter-Enc} is best interpreted as an optional residual: fixed gains are near-zero to modest, while calibrated variants are stronger but use validation labels.

\section{Analysis}
\label{sec:analysis}

\noindent\textbf{Limits of global correction.}
Our experiments show that many internal signals are descriptive rather than selective. Stability, coherence, hidden-state structure, and graph reconstruction reveal that basins exist, but they do not determine which basin is correct. A stable wrong frame can dominate the pool, while a correct challenger can remain small. This is why broad replacement policies and global rerankers often degrade raw consensus.

\noindent\textbf{Additive framing evidence.}
Framing evidence helps when it is treated as a side channel rather than a new baseline. A frame can expose a target-quantity, entity-binding, unit, or reasoning-structure disagreement between basins. However, a frame can also be coherent and wrong. \textsc{Arbiter-$\Delta$} therefore uses frames to collect additional evidence and adds that evidence to the raw support prior. The method changes the consensus decision only when the auxiliary evidence overcomes the dominant-basin prior.

\noindent\textbf{Residual role of the encoder.}
The encoder's successful role is not to verify correctness directly. Our earlier basin encoder ranked risk and abstention well, but review-based correction was negative. The revised role is narrower: identify when framed, panel, or guided evidence is noisy; reduce degraded flips; and provide a clipped residual in the same log-evidence space as \textsc{Arbiter-$\Delta$}. This keeps hidden-state structure useful without allowing it to override consensus alone.

\noindent\textbf{Diagnostic role of basin graphs.}
The graph branch remains valuable as an explanatory object even though the router was negative as a selector. A Basin Story Graph can visualize how sampled trajectories split into basins, which frames conflict, and which evidence sources support each basin. We treat this as diagnostic structure rather than as the main selection mechanism.

\section{Limitations}
\label{sec:limitations}

\textsc{Arbiter-$\Delta$} uses additional evidence beyond the raw consensus pool. We therefore report the evidence budget and avoid matched-compute claims against raw consensus. All auxiliary evidence comes from the same frozen base model but still requires additional generation.
Baseline values differ across run families because prompts, seeds, dataset adapters, and scripts differ, and even nominally identical runs can vary due to nondeterminism in batched inference (e.g., vLLM scheduling). We therefore compare each method only to the raw consensus baseline from the same run family. Oracle numbers are diagnostic ceilings, not achievable results without gold labels.
The log-linear source-pooling view of Eq.~(M2) is not a calibrated Bayesian posterior over all auxiliary samples. It treats each evidence source as a bounded empirical opinion; this avoids letting larger same-model auxiliary budgets automatically dominate the raw consensus term. However, framed-pool, panel, and guided evidence all come from the same frozen model and can share correlated mistakes, which can lead to double-counting. The current method mitigates this with reliability shrinkage and a conservative sign rule, but direct dependence modeling remains future work.
The calibrated encoder variant uses labeled validation data to tune the residual rule, so it does not satisfy the strict zero-external-information setting. The fixed encoder variant remains compatible with zero external information but produces smaller and sometimes near-zero gains.
The experiments cover three frozen instruction-tuned model families and three math benchmarks. The design is model-size agnostic, but frontier-scale models, non-math tasks, and tasks with harder answer equivalence remain future work. Results can also depend on answer extraction and equivalence rules, especially for symbolic math. Appendix~\ref{app:three_seed_replication} reports standard deviations over three random seeds for a closely related Qwen3-4B GSM8K source-set diagnostic; broader seed replications for the fixed main policy remain future work.
\section{Conclusion}
\label{sec:conclusion}

Raw consensus is a strong dominant-basin estimator but leaves a wrong-majority failure case: the correct answer can appear in the pool yet lose to a larger wrong basin. \textsc{Arbiter} addresses this as basin arbitration under zero external information. \textsc{Arbiter-$\Delta$} keeps consensus as the prior and adds same-model evidence only when it supports an alternative. Across the $3\times3$ matrix and tested seeds, it consistently exceeds raw consensus, with accuracy gain in eight cells. The broader lesson: structure is not truth. Reliable recovery requires conservative additive evidence and explicit accounting of recovered and degraded cases.

\bibliographystyle{plainnat}
\bibliography{references}

\begin{thebibliography}{31}
\providecommand{\natexlab}[1]{#1}
\providecommand{\url}[1]{\texttt{#1}}
\expandafter\ifx\csname urlstyle\endcsname\relax
  \providecommand{\doi}[1]{doi: #1}\else
  \providecommand{\doi}{doi: \begingroup \urlstyle{rm}\Url}\fi

\bibitem[Abdin et~al.(2024)Abdin, Aneja, Behl, Bubeck, Eldan, Gunasekar, Harrison, Hewett, Javaheripi, Kauffmann, Lee, Lee, Li, Liu, Mendes, Nguyen, Price, de~Rosa, Saarikivi, Salim, Shah, Wang, Ward, Wu, Yu, Zhang, and Zhang]{abdin2024phi4}
Marah Abdin, Jyoti Aneja, Harkirat Behl, S{\'e}bastien Bubeck, Ronen Eldan, Suriya Gunasekar, Michael Harrison, Russell~J. Hewett, Mojan Javaheripi, Piero Kauffmann, James~R. Lee, Yin~Tat Lee, Yuanzhi Li, Weishung Liu, Caio C.~T. Mendes, Anh Nguyen, Eric Price, Gustavo de~Rosa, Olli Saarikivi, Adil Salim, Shital Shah, Xin Wang, Rachel Ward, Yue Wu, Dingli Yu, Cyril Zhang, and Yi~Zhang.
\newblock {Phi-4} technical report.
\newblock \emph{arXiv preprint arXiv:2412.08905}, 2024.
\newblock URL \url{https://arxiv.org/abs/2412.08905}.

\bibitem[Chen et~al.(2026)Chen, Cheng, Han, and Keselj]{chen2026instability}
Jinkun Chen, Fengxiang Cheng, Sijia Han, and Vlado Keselj.
\newblock {``I} may not have articulated myself clearly'': Diagnosing dynamic instability in {LLM} reasoning at inference time.
\newblock \emph{arXiv preprint arXiv:2602.02863}, 2026.
\newblock URL \url{https://arxiv.org/abs/2602.02863}.

\bibitem[Chen et~al.(2023)Chen, Aksitov, Alon, Ren, Xiao, Yin, Prakash, Sutton, Wang, and Zhou]{chen2023usc}
Xinyun Chen, Renat Aksitov, Uri Alon, Jie Ren, Kefan Xiao, Pengcheng Yin, Sushant Prakash, Charles Sutton, Xuezhi Wang, and Denny Zhou.
\newblock Universal self-consistency for large language model generation.
\newblock \emph{arXiv preprint arXiv:2311.17311}, 2023.
\newblock URL \url{https://arxiv.org/abs/2311.17311}.

\bibitem[Cobbe et~al.(2021)Cobbe, Kosaraju, Bavarian, Chen, Jun, Kaiser, Plappert, Tworek, Hilton, Nakano, Hesse, and Schulman]{cobbe2021gsm8k}
Karl Cobbe, Vineet Kosaraju, Mohammad Bavarian, Mark Chen, Heewoo Jun, Lukasz Kaiser, Matthias Plappert, Jerry Tworek, Jacob Hilton, Reiichiro Nakano, Christopher Hesse, and John Schulman.
\newblock Training verifiers to solve math word problems.
\newblock \emph{arXiv preprint arXiv:2110.14168}, 2021.
\newblock URL \url{https://arxiv.org/abs/2110.14168}.

\bibitem[Ghasemabadi and Niu(2025)]{ghasemabadi2024gnosis}
Amirhosein Ghasemabadi and Di~Niu.
\newblock Can {LLMs} predict their own failures? self-awareness via internal circuits.
\newblock \emph{arXiv preprint arXiv:2512.20578}, 2025.
\newblock URL \url{https://arxiv.org/abs/2512.20578}.

\bibitem[Grattafiori et~al.(2024)Grattafiori, Dubey, Jauhri, et~al.]{dubey2024llama3}
Aaron Grattafiori, Abhimanyu Dubey, Abhinav Jauhri, et~al.
\newblock The {Llama} 3 herd of models.
\newblock \emph{arXiv preprint arXiv:2407.21783}, 2024.
\newblock URL \url{https://arxiv.org/abs/2407.21783}.

\bibitem[Han et~al.(2025)Han, Wang, Zhao, Li, Jiang, Jiang, Liang, Lin, Zhou, Sun, Yu, and Xiao]{han2025stitch}
Jinyi Han, Xinyi Wang, Haiquan Zhao, Tingyun Li, Zishang Jiang, Sihang Jiang, Jiaqing Liang, Xin Lin, Weikang Zhou, Zeye Sun, Fei Yu, and Yanghua Xiao.
\newblock A stitch in time saves nine: Proactive self-refinement for language models.
\newblock \emph{arXiv preprint arXiv:2508.12903}, 2025.
\newblock URL \url{https://arxiv.org/abs/2508.12903}.

\bibitem[Hendrycks et~al.(2021{\natexlab{a}})Hendrycks, Burns, Basart, Zou, Mazeika, Song, and Steinhardt]{hendrycks2021mmlu}
Dan Hendrycks, Collin Burns, Steven Basart, Andy Zou, Mantas Mazeika, Dawn Song, and Jacob Steinhardt.
\newblock Measuring massive multitask language understanding.
\newblock In \emph{International Conference on Learning Representations}, 2021{\natexlab{a}}.
\newblock URL \url{https://arxiv.org/abs/2009.03300}.

\bibitem[Hendrycks et~al.(2021{\natexlab{b}})Hendrycks, Burns, Kadavath, Arora, Basart, Tang, Song, and Steinhardt]{hendrycks2021math}
Dan Hendrycks, Collin Burns, Saurav Kadavath, Akul Arora, Steven Basart, Eric Tang, Dawn Song, and Jacob Steinhardt.
\newblock Measuring mathematical problem solving with the {MATH} dataset.
\newblock In \emph{Advances in Neural Information Processing Systems}, 2021{\natexlab{b}}.
\newblock URL \url{https://arxiv.org/abs/2103.03874}.

\bibitem[Huang et~al.(2024)Huang, Chen, Mishra, Zheng, Yu, Song, and Zhou]{huang2024large}
Jie Huang, Xinyun Chen, Swaroop Mishra, Huaixiu~Steven Zheng, Adams~Wei Yu, Xinying Song, and Denny Zhou.
\newblock Large language models cannot self-correct reasoning yet.
\newblock In \emph{International Conference on Learning Representations}, 2024.
\newblock URL \url{https://arxiv.org/abs/2310.01798}.

\bibitem[Knappe et~al.(2024)Knappe, Li, Chauhan, Chhua, Zhu, and O'Brien]{semanticselfconsistency2024}
Tim Knappe, Ryan Li, Ayush Chauhan, Kaylee Chhua, Kevin Zhu, and Sean O'Brien.
\newblock Semantic self-consistency: Enhancing language model reasoning via semantic weighting, 2024.
\newblock URL \url{https://arxiv.org/abs/2410.07839}.

\bibitem[Kwon et~al.(2023)Kwon, Li, Zhuang, Sheng, Zheng, Yu, Gonzalez, Zhang, and Stoica]{kwon2023vllm}
Woosuk Kwon, Zhuohan Li, Siyuan Zhuang, Ying Sheng, Lianmin Zheng, Cody~Hao Yu, Joseph~E. Gonzalez, Hao Zhang, and Ion Stoica.
\newblock Efficient memory management for large language model serving with {PagedAttention}.
\newblock In \emph{Proceedings of the ACM SIGOPS 29th Symposium on Operating Systems Principles}, 2023.
\newblock URL \url{https://arxiv.org/abs/2309.06180}.

\bibitem[Lhoest et~al.(2021)Lhoest, del Moral, Jernite, Thakur, von Platen, Patil, Chaumond, Drame, Plu, Tunstall, Davison, {\v{S}}a{\v{s}}ko, Chhablani, Malik, Brandeis, Scao, Sanh, Xu, Patry, McMillan-Major, Schmid, Gugger, Delangue, Matussi{\`e}re, Debut, Bekman, Cistac, Goehringer, Mustar, Lagunas, Rush, and Wolf]{lhoest2021datasets}
Quentin Lhoest, Albert~Villanova del Moral, Yacine Jernite, Abhishek Thakur, Patrick von Platen, Suraj Patil, Julien Chaumond, Mariama Drame, Julien Plu, Lewis Tunstall, Joe Davison, Mario {\v{S}}a{\v{s}}ko, Gunjan Chhablani, Bhavitvya Malik, Simon Brandeis, Teven~Le Scao, Victor Sanh, Canwen Xu, Nicolas Patry, Angelina McMillan-Major, Philipp Schmid, Sylvain Gugger, Cl{\'e}ment Delangue, Th{\'e}o Matussi{\`e}re, Lysandre Debut, Stas Bekman, Pierric Cistac, Thibault Goehringer, Victor Mustar, Fran{\c{c}}ois Lagunas, Alexander~M. Rush, and Thomas Wolf.
\newblock Datasets: A community library for natural language processing.
\newblock In \emph{Proceedings of the 2021 Conference on Empirical Methods in Natural Language Processing: System Demonstrations}, pages 175--184, Online and Punta Cana, Dominican Republic, 2021. Association for Computational Linguistics.
\newblock \doi{10.18653/v1/2021.emnlp-demo.21}.
\newblock URL \url{https://aclanthology.org/2021.emnlp-demo.21/}.

\bibitem[Li et~al.(2024)Li, Cui, Zhao, Kong, and Bi]{li2024gsmplus}
Qintong Li, Leyang Cui, Xueliang Zhao, Lingpeng Kong, and Wei Bi.
\newblock {GSM}-plus: A comprehensive benchmark for evaluating the robustness of {LLM}s as mathematical problem solvers.
\newblock In \emph{Proceedings of the 62nd Annual Meeting of the Association for Computational Linguistics (Volume 1: Long Papers)}, pages 2961--2984, Bangkok, Thailand, 2024. Association for Computational Linguistics.
\newblock \doi{10.18653/v1/2024.acl-long.163}.
\newblock URL \url{https://aclanthology.org/2024.acl-long.163/}.

\bibitem[Li et~al.(2025)Li, Callanan, Ghassel, and Zhu]{entropy2025branching}
Xianzhi Li, Ethan Callanan, Abdellah Ghassel, and Xiaodan Zhu.
\newblock Entropy-gated branching for efficient test-time reasoning.
\newblock \emph{arXiv preprint arXiv:2503.21961}, 2025.
\newblock URL \url{https://arxiv.org/abs/2503.21961}.

\bibitem[Liang et~al.(2025)Liang, Li, Zhou, Song, Yu, Du, Mi, and Yu]{clue2025}
Zhenwen Liang, Ruosen Li, Yujun Zhou, Linfeng Song, Dian Yu, Xinya Du, Haitao Mi, and Dong Yu.
\newblock {CLUE}: Non-parametric verification from experience via hidden-state clustering, 2025.
\newblock URL \url{https://arxiv.org/abs/2510.01591}.

\bibitem[Liang et~al.(2026)Liang, Huang, Wang, and Zhang]{liang2026step}
Zhixiang Liang, Beichen Huang, Zheng Wang, and Minjia Zhang.
\newblock Hidden states as early signals: Step-level trace evaluation and pruning for efficient test-time scaling.
\newblock \emph{arXiv preprint arXiv:2601.09093}, 2026.
\newblock URL \url{https://arxiv.org/abs/2601.09093}.

\bibitem[Lightman et~al.(2024)Lightman, Kosaraju, Burda, Edwards, Baker, Lee, Leike, Schulman, Sutskever, and Cobbe]{lightman2023letsverify}
Hunter Lightman, Vineet Kosaraju, Yura Burda, Harri Edwards, Bowen Baker, Teddy Lee, Jan Leike, John Schulman, Ilya Sutskever, and Karl Cobbe.
\newblock Let's verify step by step.
\newblock In \emph{International Conference on Learning Representations}, 2024.
\newblock URL \url{https://arxiv.org/abs/2305.20050}.

\bibitem[Madaan et~al.(2023)Madaan, Tandon, Gupta, Hallinan, Gao, Wiegreffe, Alon, Dziri, Prabhumoye, Yang, Gupta, Majumder, Hermann, Welleck, Yazdanbakhsh, and Clark]{madaan2023selfrefine}
Aman Madaan, Niket Tandon, Prakhar Gupta, Skyler Hallinan, Luyu Gao, Sarah Wiegreffe, Uri Alon, Nouha Dziri, Shrimai Prabhumoye, Yiming Yang, Shashank Gupta, Bodhisattwa~Prasad Majumder, Katherine Hermann, Sean Welleck, Amir Yazdanbakhsh, and Peter Clark.
\newblock Self-refine: Iterative refinement with self-feedback.
\newblock In \emph{Advances in Neural Information Processing Systems}, 2023.
\newblock URL \url{https://proceedings.neurips.cc/paper_files/paper/2023/hash/91edff07232fb1b55a505a9e9f6c0ff3-Abstract-Conference.html}.

\bibitem[Mirzadeh et~al.(2024)Mirzadeh, Alizadeh, Shahrokhi, Tuzel, Bengio, and Farajtabar]{mirzadeh2024gsmsymbolic}
Iman Mirzadeh, Keivan Alizadeh, Hooman Shahrokhi, Oncel Tuzel, Samy Bengio, and Mehrdad Farajtabar.
\newblock {GSM-Symbolic}: Understanding the limitations of mathematical reasoning in large language models.
\newblock \emph{arXiv preprint arXiv:2410.05229}, 2024.
\newblock URL \url{https://arxiv.org/abs/2410.05229}.

\bibitem[Oh and Lee(2025)]{latentselfconsistency2025}
Jungsuk Oh and Jay-Yoon Lee.
\newblock Latent self-consistency for reliable majority-set selection in short- and long-answer reasoning, 2025.
\newblock URL \url{https://arxiv.org/abs/2508.18395}.

\bibitem[Paszke et~al.(2019)Paszke, Gross, Massa, Lerer, Bradbury, Chanan, Killeen, Lin, Gimelshein, Antiga, Desmaison, K\"opf, Yang, DeVito, Raison, Tejani, Chilamkurthy, Steiner, Fang, Bai, and Chintala]{paszke2019pytorch}
Adam Paszke, Sam Gross, Francisco Massa, Adam Lerer, James Bradbury, Gregory Chanan, Trevor Killeen, Zeming Lin, Natalia Gimelshein, Luca Antiga, Alban Desmaison, Andreas K\"opf, Edward Yang, Zachary DeVito, Martin Raison, Alykhan Tejani, Sasank Chilamkurthy, Benoit Steiner, Lu~Fang, Junjie Bai, and Soumith Chintala.
\newblock {PyTorch}: An imperative style, high-performance deep learning library.
\newblock In \emph{Advances in Neural Information Processing Systems}, 2019.
\newblock URL \url{https://arxiv.org/abs/1912.01703}.

\bibitem[Shi et~al.(2026)Shi, Zhu, Shi, Zhang, Wang, and Miao]{shi2026stir}
Zhenning Shi, Yijia Zhu, Junhan Shi, Xun Zhang, Lei Wang, and Congcong Miao.
\newblock Internalizing {LLM} reasoning via discovery and replay of latent actions.
\newblock \emph{arXiv preprint arXiv:2602.04925}, 2026.
\newblock URL \url{https://arxiv.org/abs/2602.04925}.

\bibitem[Shinn et~al.(2023)Shinn, Cassano, Gopinath, Narasimhan, and Yao]{shinn2023reflexion}
Noah Shinn, Federico Cassano, Ashwin Gopinath, Karthik~R. Narasimhan, and Shunyu Yao.
\newblock Reflexion: Language agents with verbal reinforcement learning.
\newblock In \emph{Advances in Neural Information Processing Systems}, 2023.
\newblock URL \url{https://proceedings.neurips.cc/paper_files/paper/2023/hash/1b44b878bb782e6954cd888628510e90-Abstract-Conference.html}.

\bibitem[Snell et~al.(2025)Snell, Lee, Xu, and Kumar]{snell2024scaling}
Charlie~Victor Snell, Jaehoon Lee, Kelvin Xu, and Aviral Kumar.
\newblock Scaling {LLM} test-time compute optimally can be more effective than scaling parameters for reasoning.
\newblock In \emph{International Conference on Learning Representations}, 2025.
\newblock URL \url{https://openreview.net/forum?id=4FWAwZtd2n}.

\bibitem[Vasudev et~al.(2026)Vasudev, Russak, Bikel, and Alshikh]{vasudev2026failure}
Rakshith Vasudev, Melisa Russak, Dan Bikel, and Waseem Alshikh.
\newblock Accurate failure prediction in agents does not imply effective failure prevention.
\newblock \emph{arXiv preprint arXiv:2602.03338}, 2026.
\newblock URL \url{https://arxiv.org/abs/2602.03338}.

\bibitem[Wang et~al.(2023)Wang, Wei, Schuurmans, Le, Chi, Narang, Chowdhery, and Zhou]{wang2022selfconsistency}
Xuezhi Wang, Jason Wei, Dale Schuurmans, Quoc~V. Le, Ed~H. Chi, Sharan Narang, Aakanksha Chowdhery, and Denny Zhou.
\newblock Self-consistency improves chain of thought reasoning in language models.
\newblock In \emph{International Conference on Learning Representations}, 2023.
\newblock URL \url{https://arxiv.org/abs/2203.11171}.

\bibitem[Wei et~al.(2022)Wei, Wang, Schuurmans, Bosma, Ichter, Xia, Chi, Le, and Zhou]{wei2022cot}
Jason Wei, Xuezhi Wang, Dale Schuurmans, Maarten Bosma, Brian Ichter, Fei Xia, Ed~H. Chi, Quoc~V. Le, and Denny Zhou.
\newblock Chain-of-thought prompting elicits reasoning in large language models.
\newblock In \emph{Advances in Neural Information Processing Systems}, 2022.
\newblock URL \url{https://arxiv.org/abs/2201.11903}.

\bibitem[Wolf et~al.(2020)Wolf, Debut, Sanh, Chaumond, Delangue, Moi, Cistac, Rault, Louf, Funtowicz, Davison, Shleifer, von Platen, Ma, Jernite, Plu, Xu, Scao, Gugger, Drame, Lhoest, and Rush]{wolf2020transformers}
Thomas Wolf, Lysandre Debut, Victor Sanh, Julien Chaumond, Clement Delangue, Anthony Moi, Pierric Cistac, Tim Rault, R{\'e}mi Louf, Morgan Funtowicz, Joe Davison, Sam Shleifer, Patrick von Platen, Clara Ma, Yacine Jernite, Julien Plu, Canwen Xu, Teven~Le Scao, Sylvain Gugger, Mariama Drame, Quentin Lhoest, and Alexander~M. Rush.
\newblock Transformers: State-of-the-art natural language processing.
\newblock In \emph{Proceedings of the 2020 Conference on Empirical Methods in Natural Language Processing: System Demonstrations}, pages 38--45, Online, 2020. Association for Computational Linguistics.
\newblock \doi{10.18653/v1/2020.emnlp-demos.6}.
\newblock URL \url{https://aclanthology.org/2020.emnlp-demos.6/}.

\bibitem[Yang et~al.(2025)Yang, Li, Yang, et~al.]{yang2025qwen3}
An~Yang, Anfeng Li, Baosong Yang, et~al.
\newblock Qwen3 technical report.
\newblock \emph{arXiv preprint arXiv:2505.09388}, 2025.
\newblock URL \url{https://arxiv.org/abs/2505.09388}.

\bibitem[Zhang et~al.(2024)Zhang, Khalifa, Logeswaran, Kim, Lee, Lee, and Wang]{zhang2024smalllm}
Yunxiang Zhang, Muhammad Khalifa, Lajanugen Logeswaran, Jaekyeom Kim, Moontae Lee, Honglak Lee, and Lu~Wang.
\newblock Small language models need strong verifiers to self-correct reasoning.
\newblock In \emph{Findings of the Association for Computational Linguistics: ACL 2024}, pages 15637--15653, Bangkok, Thailand, 2024. Association for Computational Linguistics.
\newblock \doi{10.18653/v1/2024.findings-acl.924}.
\newblock URL \url{https://aclanthology.org/2024.findings-acl.924/}.

\end{thebibliography}

\appendix

\clearpage
\section{Experimental program overview table}
\label{app:program_overview}

\renewcommand{\arraystretch}{1.5}
\begin{table}[!htbp]
\centering
\caption{Empirical program summary across all major experiment families.
``Acc.\ gain range'' is the observed range of accuracy change over the
in-family raw consensus baseline (or over greedy for the consensus row). Unless a row states otherwise, representative numbers are from Qwen3-4B on the GSM8K test split. Most
correction routes fail; the positive results concentrate in additive basin
arbitration.}
\label{tab:experiment_program}
\footnotesize
\setlength{\tabcolsep}{4pt}
\begin{tabular}{%
>{\raggedright\arraybackslash}p{1.8cm}
>{\raggedright\arraybackslash}p{3.5cm}
>{\raggedright\arraybackslash}p{2.1cm}
>{\raggedright\arraybackslash}p{3.7cm}
>{\raggedright\arraybackslash}p{1.3cm}
}
\toprule
\textbf{Family} & \textbf{Concise method}
& \textbf{Acc.\ gain range (\%)} & \textbf{Main numeric outcome}
& \textbf{Status} \\
\midrule
Raw consensus
& $K{=}24$ sampled solutions, answer-cluster majority.
& $-2.2$ to $+15.2$ vs.\ greedy
& Improves over greedy in 7/9 evaluated cells, matches in one, and is lower in one.
& Clean baseline \\
Hidden-state geometry
& Coherence/divergence/branch-window scoring of trajectories.
& $\sim 0$ to $+0.2$
& Historical hidden-state rerank $94.92\%$; local-divergence slice $\approx +2/162$.
& Weak / fragile \\
Text-space steering
& Donor-fragment continuation; steerability ranking.
& $\sim 0$ ($+1/162$ best)
& AUC $0.69$--$0.73$, deployment-weak.
& Mechanistic only \\
Self-review sessions
& Solve $\to$ review $\to$ final-answer consensus.
& $-1.6$ to $+0.2$ (unstable)
& Stable reruns: raw $94.69\%\to$ reviewed $93.10\%$; one earlier positive audit run is not used as stable evidence.
& Negative / non-replicable \\
Top-up vote merging
& Add extra sessions for ambiguous qids; merge votes.
& $-0.5$ to $-0.9$
& Historical guarded audit $95.22\to94.31$ after merge; raw $94.77\to94.24$.
& Negative \\
Cluster review / memos
& Review raw traces or mutual basin memos.
& $-6$ to $-12$ vs.\ slice cons.
& Raw-trace review $70.87\%$ vs.\ $76.96\%$; answer-memo review $63$--$66.5\%$ vs.\ $74.71\%$.
& Negative \\
Answer-only hard-slice review
& Two-turn solve/review with candidate answers visible.
& $+5.9$ on hard slice
& Hard slice $64.71\%\to70.59\%$.
& Slice-only positive \\
Basin-principle judging
& Pairwise basin-card tournament.
& $-5$ to $-22$ net
& Full slice: 10 over.\ / 1 rec.\ / 6 deg.; no-guard: $-22$ net.
& Negative \\
Our original basin encoder (old target)
& Trajectory encoder for difficulty + review filtering.
& Abstention: kept-set $+4.6$; review filtering: $-46$ net
& 50\% coverage $\to 98.79\%$ kept-set accuracy; review filtering produced only 5/362 rescues.
& Risk-only positive \\
Cluster GNN/router
& Graph router over basin structure.
& Net negative at all tested budgets
& Qwen3-4B GSM8K cons.\ $94.54\%$, top-2 oracle $97.27\%$; router scores $\not\propto$ truth.
& Negative selector \\
Framing-first replacement
& Use framing-prompt output as replacement consensus.
& $-1.1$ to $+0.4$ across saved replacement variants
& Flat framed replacement $94.84\%$ but style-balanced variants fall to $94.31\%$ and $93.71\%$.
& Fragile \\
Framing additive evidence
& Basin frames as side evidence over consensus.
& $+0.3$ ($+4$ net)
& Qwen GSM8K main framed+guided $94.54\%\to94.84\%$; 9 rec.\ / 5 deg.
& Clean positive \\
\textsc{Arbiter-$\Delta$}
& Additive log-linear evidence over raw plus the fixed framed+guided source set.
& $0.0$ to $+3.0$
& 3$\times$3 matrix: 8 accuracy-positive, 1 neutral; count-level net non-negative in all cells.
& \textbf{Main clean result} \\
\textsc{Arbiter-Enc}
& Bounded trajectory residual on top of $\Delta$.
& Fixed: $-0.1$ to $+0.5$; Cal.: $+0.8$ to $+1.5$
& Larger gains where $\Delta$ leaves more residual headroom.
& Stronger residual variant \\
\bottomrule
\end{tabular}
\end{table}
\renewcommand{\arraystretch}{1.0}

\clearpage
\section{Symbol reference for setup and method}
\label{app:symbol_reference}

\begin{table}[!htbp]
\centering
\caption{Symbols used in the problem setup.}
\label{tab:setup_symbols}
\footnotesize
\setlength{\tabcolsep}{4pt}
\begin{tabular}{p{2.1cm}p{9.1cm}}
\toprule
\textbf{Symbol} & \textbf{Meaning} \\
\midrule
$q$ & Input question. \\
$M$ & Frozen autoregressive language model. \\
$K$ & Size of the ordinary raw sampled pool. \\
$s_i$ & The $i$th complete candidate solution. \\
$T_i$ & Token length of candidate solution $s_i$. \\
$y_{i,t}$ & Token $t$ in candidate solution $s_i$. \\
$L$ & Number of transformer layers. \\
$h_{i,t}^{(\ell)}$ & Layer-$\ell$ hidden state at token position $t$ for candidate $i$. \\
$H_i$ & Layer-by-layer, token-by-token hidden-state trajectory of candidate $s_i$. \\
$\mathrm{Ans}$ & Task-specific final-answer extractor and normalizer. \\
$a_i$ & Extracted normalized final answer of candidate $s_i$. \\
$\alpha_r$ & Normalized final answer associated with basin $r$. \\
$C_r$ & Set of candidate indices whose extracted final answer is $\alpha_r$. \\
$B_r$ & Observed answer basin: answer $\alpha_r$, support set $C_r$, and trajectories reaching it. \\
$B_1$ & Dominant basin, i.e., the largest answer cluster and raw-consensus basin. \\
$y^\star(q)$ & Gold answer used only for evaluation. \\
$\mathrm{Oracle@}k$ & Diagnostic event that the gold answer appears among the top $k$ ranked basins. \\
$\mathrm{WM}(q)$ & Wrong-majority indicator. \\
$\pi(q)$ & Arbitration policy selecting one observed basin index. \\
\bottomrule
\end{tabular}
\end{table}

\begin{table}[!htbp]
\centering
\caption{Method evidence symbols used by \textsc{Arbiter-$\Delta$}. Each count is computed per question by parsing same-model outputs and assigning the final answer to an observed basin. ``Dominant'' means the raw-consensus basin $B_1$; ``challenger'' means an alternative observed basin $B_r$.}
\label{tab:method_symbols}
\footnotesize
\setlength{\tabcolsep}{4pt}
\begin{tabular}{p{1.5cm}p{3.0cm}p{8cm}}
\toprule
\textbf{Symbol} & \textbf{Name} & \textbf{How it is computed} \\
\midrule
$b_1,b_2$ & Raw support & Number of ordinary sampled solutions in the raw pool whose final answer matches the majority basin $B_1$ or leading challenger $B_2$; equivalently $b_j=|C_j|$. \\
$f_1,f_2$ & Framed-pool support & Number of frame-oriented solves whose final answer matches $B_1$ or $B_2$. The main source family uses 24 framed solves. \\
$p_1,p_r$ & Panel support & Number of comparison-panel trials whose final answer matches $B_1$ or $B_r$, with $p_j=p_j^++p_j^-$. Panel source-set ablations use about 12 trials. \\
$g_1,g_2$ & Guided support & Number of guided re-solves whose final answer matches $B_1$ or $B_2$. The main source family uses about 4 trials. \\
$N_e^{\mathrm{att}}$ & Attempted count & Number of attempted outputs for source $e$, including invalid outputs and answers outside the compared pair. \\
$\alpha$ & Pseudocount & Fixed to $1.0$ for Laplace smoothing; it prevents zero counts from dominating a log ratio. \\
$r_f,r_g,\rho_P$ & Reliability factors & Label-free factors in $[0,1]$: attempted top-2 mass for framed/guided sources and top-pair mass times order symmetry for panel ablations. \\
\bottomrule
\end{tabular}
\end{table}

\clearpage
\section{Derivation of the additive Delta score}
\label{app:delta_derivation}

This appendix records the algebra behind Eq.~(M4). The proposition is only a score-form identity; it is not a theorem that \textsc{Arbiter-$\Delta$} strictly improves accuracy. Accuracy improvement is an empirical claim reported in Section~\ref{sec:exp_delta_main}.

\begin{proposition}[Pairwise log-linear source pooling]
Fix a question $q$, the dominant basin $B_1$, the leading challenger $B_2$, raw supports $b_1,b_2$, framed counts $f_1,f_2$, guided counts $g_1,g_2$, a pseudo-count $\alpha>0$, and reliability factors $r_f(q),r_g(q)\in[0,1]$. For $j\in\{1,2\}$ define
\[
\widetilde w_j(q)=(b_j+\alpha)(f_j+\alpha)^{r_f(q)}(g_j+\alpha)^{r_g(q)}.
\]
Then
\[
\log\frac{\widetilde w_2(q)}{\widetilde w_1(q)}
=
\log\frac{b_2+\alpha}{b_1+\alpha}
+r_f(q)\log\frac{f_2+\alpha}{f_1+\alpha}
+r_g(q)\log\frac{g_2+\alpha}{g_1+\alpha}
=\Delta_2(q).
\]
Therefore Eq.~(M3) selects the challenger exactly when its pooled support exceeds the dominant basin's pooled support.
\end{proposition}

\paragraph{Proof.}
Taking the ratio of the two definitions gives
\[
\frac{\widetilde w_2(q)}{\widetilde w_1(q)}
=
\frac{b_2+\alpha}{b_1+\alpha}
\left(\frac{f_2+\alpha}{f_1+\alpha}\right)^{r_f(q)}
\left(\frac{g_2+\alpha}{g_1+\alpha}\right)^{r_g(q)}.
\]
Taking logarithms converts the product into the additive score in Eq.~(M2). The sign rule in Eq.~(M3) is therefore equivalent to comparing $\widetilde w_2(q)$ with $\widetilde w_1(q)$.\hfill$\square$

This identity is Bayes-motivated because likelihood ratios also add in log space, but Eq.~(M2) should not be read as a calibrated Bayesian posterior over all auxiliary samples. It is a log-linear pooling rule over source-level empirical opinions. A posterior update over independent individual auxiliary trials would generally scale with the number of trials in each source; the implemented score deliberately avoids that scaling because all sources come from the same frozen model and can share correlated mistakes.

\clearpage
\section{Full empirical ledger}
\label{app:full_ledger}

Table~\ref{tab:full_empirical_ledger} records the broader project history. The
main paper focuses on the cleanest comparisons, while this appendix preserves
the full set of positive, negative, calibrated, hard-slice, and exploratory
results.

\footnotesize
\setlength{\tabcolsep}{3pt}

\begin{longtable}{%
>{\raggedright\arraybackslash}p{2cm}
>{\raggedright\arraybackslash}p{2.6cm}
>{\raggedright\arraybackslash}p{4cm}
>{\raggedright\arraybackslash}p{2.6cm}
>{\raggedright\arraybackslash}p{1.5cm}
}

\caption{Full empirical ledger. ``Clean'' means gold labels are used only after
prediction for evaluation. ``Gold-tuned'' means a configuration was selected
using target labels and is therefore reported only as an exploratory upper
bound.}
\label{tab:full_empirical_ledger} \\

\toprule
\textbf{Phase} & \textbf{Method} & \textbf{Recorded result} & \textbf{Interpretation} & \textbf{Status} \\
\midrule
\endfirsthead

\multicolumn{5}{c}%
{{\tablename\ \thetable{} -- continued from previous page}} \\
\toprule
\textbf{Phase} & \textbf{Method} & \textbf{Recorded result} & \textbf{Interpretation} & \textbf{Status} \\
\midrule
\endhead

\midrule
\multicolumn{5}{r}{{Continued on next page}} \\
\endfoot

\bottomrule
\endlastfoot

Early selector
& Consensus $\times$ robustness / resolve
& Robustness could zero out correct clusters despite high support.
& Misaligned confidence signal.
& Negative \\

raw consensus
& Raw $K{=}24$ answer clustering
& Typical GSM8K full-test accuracy $\approx94.3$--$94.7\%$; gold-in-pool about $98\%$.
& Stable backbone.
& Clean \\

Generation-time control
& Probe-delta + conservative generation-time control
& About $91.96\%$ greedy-path accuracy.
& Best generation-time branch, still below consensus.
& Clean \\

Hidden-state trajectory rerank
& Hidden-state coherence rerank
& $94.92\%$.
& Small guarded gain; fragile.
& Exploratory \\

CGAC / token-1 FFN
& Token-1 FFN activation bias and layer interventions
& Real bias effects resembled or underperformed random controls.
& Token-1 FFN not reliable.
& Negative \\

Shared-prefix branch
& Branch-window hidden-state detection and patching
& Branch divergence detectable; forcing 1--5 branch tokens and one-cell patching weak.
& Detectability $\neq$ controllability.
& Negative \\

Continuation steering
& Coherent continuation donor insertion
& Clear dose-response steering between basins.
& Strong mechanistic evidence.
& Mechanistic \\

Steerability selector
& Convert steerability into a selector score
& AUC $\approx0.69$--$0.73$; best selector about $+1/162$ qids.
& Steering signal too weak for deployment.
& Weak positive \\

Local divergence scoring
& Local divergence contrastive scoring
& About $0.7284$ on 162-qid slice; about $+2$ qids over consensus with $\sim52\%$ coverage.
& Slice signal only.
& Slice-only \\

Self-review sessions
& Solve $\rightarrow$ review $\rightarrow$ final consensus
& Stable reruns were negative: raw $94.69\%\rightarrow$ reviewed $93.10\%$. An earlier self-review run showed $94.77\%\rightarrow95.00\%$ but is treated as non-replicated.
& Broad self-review is not a stable correction method.
& Negative / unstable \\

Historical guarded sweep
& Post-hoc guarded rerank
& $95.22\%$; 7 overrides, 6 recovered, 0 degraded.
& Historical upper bound selected by gold-evaluated grid.
& Gold-tuned \\

Top-up merge
& Generate extra sessions for 593 ambiguous qids and merge votes
& Historical raw/final/guarded changed from $94.77/95.00/95.22$ to $94.24/94.54/94.31$.
& Extra votes diluted correct majorities.
& Negative \\

Raw-trace review
& Sequential review over raw traces
& Best 32k solve-at-end known-else-c1 $70.87\%$ vs $76.96\%$ consensus.
& Raw-trace review underperforms.
& Negative \\

Answer-memo review
& Mutual memos with answers
& Pair/triad reviews $\approx63$--$66.5\%$ vs $74.71\%$ consensus.
& Summaries lose decisive evidence.
& Negative \\

Answer-only review
& Question + candidate answers, two-turn solve/review
& Harder slice: $64.71\%\rightarrow70.59\%$.
& Good hard-slice solver.
& Slice-only positive \\

Local branch packets
& Branch-local packetization and answer review
& 5\% packetized review: slice $61.7\%\rightarrow66.0\%$, full-test net $+1$. Swapped-only tiny stratum reached 100\% on 3 packets.
& Useful stratum detection, low coverage.
& Slice-only \\

Basin tournament
& Basin principle cards and pairwise judging
& Full 217 multi-basin qids: 10 overrides, 1 recovered, 6 degraded, net $-5$.
& Direct basin judging unreliable.
& Negative \\

LLM basin baseline
& Basin summaries without encoder
& Without support guard: 42 overrides, 7 recovered, 29 degraded, net $-22$.
& Unsafe selector.
& Negative \\

Basin encoder Task A
& Trajectory encoder for difficulty / abstention
& 50\% coverage $\rightarrow98.79\%$ kept-set accuracy.
& Strong risk ranking.
& Clean, not full coverage \\

Basin encoder Task B
& Encoder-filtered review correction
& Only 5/362 review candidates rescued consensus errors; accept-any-review net $-46$.
& Review source is poor.
& Negative \\

Cluster GNN/router
& Graph router over basin structure
& Qwen3-4B GSM8K consensus $94.54\%$; top-2 oracle $97.27\%$; router overrides net negative.
& Structure learned, correctness not learned.
& Negative \\

Framing-first
& Framing-prompt generation as replacement
& Flat framed replacement $94.84\%$; style-balanced $94.31\%$; style+temp $93.71\%$.
& Replacement framing is fragile rather than the main deployable selector.
& Fragile \\

Framing additive
& Basin frames as side evidence
& $94.54\%\rightarrow94.84\%$; 16 overrides, 9 recovered, 5 degraded.
& Framing works additively.
& Clean positive \\

\textsc{Arbiter-$\Delta$}
& Additive log-linear evidence over raw plus the fixed framed+guided source set
& 3$\times$3 matrix: 8 accuracy-positive, 1 neutral; count-level net non-negative in all cells.
& Main clean method.
& Clean positive \\

\textsc{Arbiter-Enc}
& Framing-aware hidden-state residual
& Fixed: $-0.1$ to $+0.5$ mean extra gain. Calibrated: $+0.8$ to $+1.5$.
& Optional residual variant.
& Fixed / calibrated \\

\end{longtable}

\normalsize 

Table~\ref{tab:app_consensus_oracle} reports the consensus and oracle headroom table for the main evaluation. Method comparisons use these same in-family raw-consensus baselines for the evaluation matrix.

\begin{table}[!htbp]
\centering
\caption{Consensus and oracle headroom for the main evaluation matrix. O@k is a diagnostic same-pool oracle over the top k answer basins; count columns report additional correct examples over raw consensus.}
\label{tab:app_consensus_oracle}
\footnotesize
\setlength{\tabcolsep}{3pt}
\resizebox{\linewidth}{!}{%
\begin{tabular}{llrrrrrrrrr}
\toprule
\textbf{Model} & \textbf{Dataset} & \textbf{N}
& \textbf{Greedy} & \textbf{Cons.}
& \textbf{O@2} & \textbf{O2 Ct.}
& \textbf{O@3} & \textbf{O3 Ct.}
& \textbf{O@5} & \textbf{O5 Ct.} \\
\midrule
Llama-3.1-8B & GSM8K & 1319 & 84.69 & 91.81 & 95.68 & 51 & 96.66 & 64 & 97.42 & 74 \\
Llama-3.1-8B & MMLU-HS-Math & 270 & 63.33 & 78.52 & 91.11 & 34 & 94.81 & 44 & 98.15 & 53 \\
Llama-3.1-8B & MATH-500 & 500 & 41.60 & 51.60 & 60.80 & 46 & 63.60 & 60 & 66.20 & 73 \\
\midrule
Qwen3-4B & GSM8K & 1319 & 90.67 & 94.54 & 97.27 & 36 & 97.88 & 44 & 98.26 & 49 \\
Qwen3-4B & MMLU-HS-Math & 270 & 79.63 & 94.81 & 96.30 & 4 & 96.30 & 4 & 96.30 & 4 \\
Qwen3-4B & MATH-500 & 500 & 71.40 & 69.20 & 76.60 & 37 & 77.60 & 42 & 78.40 & 46 \\
\midrule
Phi-4 & GSM8K & 1319 & 94.92 & 96.13 & 97.95 & 24 & 98.56 & 32 & 98.79 & 35 \\
Phi-4 & MMLU-HS-Math & 270 & 87.78 & 92.59 & 98.15 & 15 & 98.89 & 17 & 98.89 & 17 \\
Phi-4 & MATH-500 & 500 & 73.00 & 73.00 & 78.20 & 26 & 80.00 & 35 & 81.40 & 42 \\
\bottomrule
\end{tabular}%
}
\end{table}

\begin{table}[!htbp]
\centering
\caption{GSM8K raw-only sampling-budget diagnostic. The $K{=}56$ rows use ordinary raw samples only, without framed, guided, panel, or encoder evidence. This diagnostic is included to show that the GSM8K gains are not explained by simply drawing more ordinary samples, but it is not a strict compute-matched full-matrix baseline.}
\label{tab:gsm8k_k56_raw}
\small
\begin{tabular}{lrrr}
\toprule
\textbf{Model} & \textbf{Raw cons. $K{=}24$} & \textbf{Raw cons. $K{=}56$} & \textbf{Change (pp)} \\
\midrule
Qwen3-4B & 94.54 & 94.39 & $-0.15$ \\
Llama-3.1-8B & 91.81 & 91.81 & $+0.00$ \\
Phi-4 & 96.13 & 95.75 & $-0.38$ \\
\bottomrule
\end{tabular}
\end{table}

\clearpage
\section{Self-review and top-up results}
\label{app:self_review_topup}

\begin{table}[!htbp]
\centering
\caption{Self-review audit. The stable conclusion is negative: broad review reduces accuracy and increases fragmentation. The historical self-review improvement is listed for completeness but is not used as the main stable evidence.}
\label{tab:app_self_review_audit}
\small
\begin{tabular}{lrrl}
\toprule
\textbf{Run family} & \textbf{Before review} & \textbf{After review} & \textbf{Status} \\
\midrule
Later broad review rerun & 94.69\% & 93.10\% & Stable negative signal \\
Earlier self-review session run & 94.77\% & 95.00\% & Non-replicated; fragments clusters \\
\bottomrule
\end{tabular}
\end{table}

\begin{table}[!htbp]
\centering
\caption{Earlier self-review session run details on GSM8K. This run is kept as an audit record, not as the stable self-review claim.}
\label{tab:app_self_review_hist}
\small
\begin{tabular}{lr}
\toprule
\textbf{Quantity} & \textbf{Value} \\
\midrule
Questions & 1319 \\
Sessions per qid & 25 \\
Greedy anchor base accuracy & 91.36\% \\
Greedy anchor final accuracy & 87.95\% \\
Base consensus accuracy & 94.77\% \\
Reviewed consensus accuracy in this run & 95.00\% \\
Mean raw clusters/qid & 2.553 \\
Mean final clusters/qid & 3.014 \\
Qids with 3+ raw clusters & 484 \\
Qids with 3+ final clusters & 657 \\
\bottomrule
\end{tabular}
\end{table}

\begin{table}[!htbp]
\centering
\caption{Historical post-hoc guarded sweep. This is gold-tuned unless the selected
configuration is frozen on a separate validation split and retested.}
\label{tab:app_guarded_historical}
\small
\begin{tabular}{lr}
\toprule
\textbf{Quantity} & \textbf{Value} \\
\midrule
Raw consensus & 94.77\% \\
Reviewed consensus in this run & 95.00\% \\
Best guarded rerank accuracy & 95.22\% \\
Overrides & 7 \\
Recovered & 6 \\
Degraded & 0 \\
Net & +6 \\
Gain vs raw & +0.45\% \\
Gain vs reviewed consensus in this run & +0.23\% \\
\bottomrule
\end{tabular}
\end{table}

\begin{table}[!htbp]
\centering
\caption{top-up targeted top-up merge. Extra sessions were generated for 593
ambiguous qids and merged into the vote.}
\label{tab:app_top-up_topup}
\small
\begin{tabular}{lrr}
\toprule
\textbf{Metric} & \textbf{Before merge} & \textbf{After merge} \\
\midrule
Raw consensus & 94.77\% & 94.24\% \\
Final consensus & 95.00\% & 94.54\% \\
Guarded accuracy & 95.22\% & 94.31\% \\
\midrule
Changed qids, raw/final/guarded & -- & 17 / 15 / 19 \\
Recovered, raw/final/guarded & -- & 3 / 2 / 2 \\
Degraded, raw/final/guarded & -- & 10 / 8 / 14 \\
\bottomrule
\end{tabular}
\end{table}

\clearpage
\section{Cluster review and answer-only review details}
\label{app:review_details}

\begin{table}[!htbp]
\centering
\caption{Cluster review and answer-review results. These experiments show that
explicit cluster judging and abstract memos usually underperform consensus,
while answer-only review can help on selected hard slices.}
\label{tab:app_review_details}
\footnotesize
\setlength{\tabcolsep}{3pt}
\begin{tabular}{%
>{\raggedright\arraybackslash}p{3cm}
>{\raggedright\arraybackslash}p{4.5cm}
>{\raggedright\arraybackslash}p{5.5cm}}
\toprule
\textbf{Experiment} & \textbf{Setup} & \textbf{Recorded result} \\
\midrule
Raw-pool cluster build
& Raw-consensus-style pool and disagreement slice.
& Base pool: 1319 qids, 31656 generations, 100\% parseable, mean unique
clusters/qid 2.347, median 2, selected qids under broad 2-cluster rule 795,
overall consensus 94.54\%. \\
Raw-trace review: explicit cluster choice
& Explicitly choose the correct cluster from raw traces.
& On relaxed 242-qid slice: consensus 78.10\%, candidate-basin oracle 90.08\%.
Pick-only variants ranged from 42.98\% to 61.57\%. \\
Raw-trace review: solve at end
& Review raw traces, then solve fresh.
& On 230-qid relaxed slice: consensus 76.96\%, oracle 91.30\%. Best 32k
candidate-cluster solve-at-end with consensus fallback: 70.87\%. \\
Hidden-mutual trajectory review
& Summarize hidden mutual trajectories without answer leakage.
& Hidden-mutual extraction parse rate 95.98\%; scrub parse rate 79.39\%; pair/triad
with consensus fallback about 68.7--70.0\% vs 76.96\% consensus. \\
Answer-memo review
& Pair/triad review of cluster memos with answers included.
& 170-qid slice: consensus 74.71\%, top-2 oracle 88.82\%, top-3 oracle
91.18\%. Pair/triad review results about 63.2--66.5\%. \\
Answer-only review, broad hard slice
& Question plus candidate numeric answers; two-turn solve/review.
& 169 qids: consensus 69.82\%, shown-answer oracle 87.57\%, turn-2 majority
71.60\%, turn-2 with consensus fallback 72.19\%. \\
Answer-only review, harder slice
& Top candidates with stronger challenger support.
& 119 qids: consensus 64.71\%, shown-answer oracle 87.39\%, turn-2 majority
68.91\%, turn-2 with consensus fallback 70.59\%. \\
\bottomrule
\end{tabular}
\end{table}

\clearpage
\section{Local-branch packetization}
\label{app:local_branch}

\begin{table}[!htbp]
\centering
\caption{Local-branch packetized review. Results are useful mainly as stratum
detection, not as a full-coverage method.}
\label{tab:app_local_branch}
\small
\begin{tabular}{rrrrrr}
\toprule
\textbf{Budget} & \textbf{Packets} & \textbf{Slice cons.}
& \textbf{Slice final} & \textbf{Oracle} & \textbf{Full net} \\
\midrule
1\% & 9 & 66.7 & 55.6 & 88.9 & $-2$ \\
2\% & 19 & 63.2 & 63.2 & 78.9 & $-1$ \\
5\% & 47 & 61.7 & 66.0 & 85.1 & $+1$ \\
\bottomrule
\end{tabular}
\end{table}

\begin{table}[!htbp]
\centering
\caption{Swapped-only local-branch stratum. This tiny stratum carried
concentrated rescue signal.}
\label{tab:app_local_swapped}
\small
\begin{tabular}{lrrrr}
\toprule
\textbf{Condition} & \textbf{Packets} & \textbf{Slice cons.}
& \textbf{Slice final} & \textbf{Net} \\
\midrule
plain candidate-basin review, swapped only & 3 & 33.3 & 100.0 & +2 \\
observed-only context, swapped only & 3 & -- & 100.0 & +1 \\
expansions-only context, swapped only & 3 & -- & 100.0 & +1 \\
\bottomrule
\end{tabular}
\end{table}

\clearpage
\section{Basin encoder diagnostics}
\label{app:encoder_diagnostics}

Our original basin encoder was useful for risk ranking but not for correction.
This motivates the revised \textsc{Arbiter-Enc} design, where the encoder is
used only as a bounded trajectory residual.

\begin{table}[!htbp]
\centering
\caption{Basin encoder training and data summary.}
\label{tab:app_encoder_data}
\small
\begin{tabular}{lr}
\toprule
\textbf{Quantity} & \textbf{Value} \\
\midrule
Total qids & 1319 \\
Qids with at least two stable basins & 217 \\
Basin histogram & $\{1:1102,\,2:142,\,3:75\}$ \\
Feature shards & 21 \\
Feature dimension & 384 \\
Best checkpoint & epoch 4 \\
Baseline accuracy for Task A & 94.24\% \\
Pearson confidence vs correctness & 0.168 \\
Spearman confidence vs correctness & 0.198 \\
\bottomrule
\end{tabular}
\end{table}

\begin{table}[!htbp]
\centering
\caption{Basin encoder Task A: difficulty / abstention. Kept-set accuracy is reported after abstaining on the riskiest examples; the old ambiguous ``Lift'' column is omitted.}
\label{tab:app_encoder_task_a}
\small
\begin{tabular}{rr}
\toprule
\textbf{Coverage} & \textbf{Kept-set accuracy} \\
\midrule
100\% & 94.24 \\
90\% & 94.95 \\
80\% & 95.92 \\
70\% & 97.18 \\
60\% & 98.10 \\
50\% & 98.79 \\
\bottomrule
\end{tabular}
\end{table}

\begin{table}[!htbp]
\centering
\caption{Basin encoder Task B: review filtering. The generated review
candidates are poor rescue sources.}
\label{tab:app_encoder_task_b}
\small
\begin{tabular}{lrrrrr}
\toprule
\textbf{Strategy} & \textbf{Overrides} & \textbf{Rec.}
& \textbf{Deg.} & \textbf{Net} & \textbf{Acc.} \\
\midrule
Accept any review & 60 & 0 & 46 & $-46$ & 90.75 \\
Only if differs & 173 & 4 & 150 & $-146$ & 83.17 \\
Majority agree & 7 & 0 & 5 & $-5$ & 93.86 \\
\bottomrule
\end{tabular}
\end{table}

The key ceiling result is that only 5 of 362 review candidates produced the
correct answer for a consensus error. Therefore, our old encoder is not a truth
selector. Its useful role is to rank risk and, in the new method, provide a
bounded residual over framing evidence.

\clearpage
\section{Cluster-GNN and router diagnostics}
\label{app:gnn_diagnostics}

The cluster-GNN/router branch is reported as a negative result. The graph model
learned structural objectives, but its score did not correlate with correction
quality.

\begin{table}[!htbp]
\centering
\caption{Cluster-GNN run summary on Qwen3-4B GSM8K.}
\label{tab:app_gnn_summary}
\small
\begin{tabular}{lr}
\toprule
\textbf{Quantity} & \textbf{Value} \\
\midrule
Consensus baseline & 94.54\% \\
Leading-challenger oracle & 97.27\% \\
Recoverable cases & 36 \\
Degradable cases & 680 \\
Official router overrides & Net negative at tested budgets \\
Best support-count baseline & +4 net at 0.5\% budget \\
\bottomrule
\end{tabular}
\end{table}

\begin{table}[!htbp]
\centering
\caption{Cluster-GNN diagnostic feature separation. Recoverable means consensus
is wrong and the challenger is correct. Degradable means consensus is correct
and the challenger is wrong.}
\label{tab:app_gnn_feature_diag}
\small
\begin{tabular}{lrr}
\toprule
\textbf{Feature} & \textbf{Recoverable median} & \textbf{Degradable median} \\
\midrule
GNN gain & $-0.013$ & $-0.008$ \\
Challenge probability & 0.498 & 0.506 \\
Support margin & 0.333 & 0.917 \\
Challenger count & 6.0 & 1.0 \\
$E1_{\mathrm{self}}$ & 52.7 & 52.0 \\
\bottomrule
\end{tabular}
\end{table}

Simple support geometry separated recoverable and degradable cases better than
the learned GNN score. This is why the graph/router branch motivates the
framing pivot rather than serving as the main positive selector.

\clearpage
\section{Detailed framing and delta results}
\label{app:delta_details}

\subsection{Replacement versus additive framing}

\begin{table}[!htbp]
\centering
\caption{Replacement versus additive framing evidence on Qwen3-4B GSM8K. Framing information is safest when accumulated as side evidence; direct replacement can be positive in a narrow flat variant but is fragile under style/temperature changes and unsafe for panel/guided replacement.}
\label{tab:app_replacement_additive}
\small
\setlength{\tabcolsep}{3pt}
\begin{tabular}{>{\raggedright\arraybackslash}p{2.5cm}rrrrrp{6.5cm}}
\toprule
\textbf{Policy} & \textbf{Acc.} & \textbf{Over.}
& \textbf{Rec.} & \textbf{Deg.} & \textbf{Net} & \textbf{Interpretation} \\
\midrule
Raw comparison consensus & 94.54 & 0 & 0 & 0 & 0
& In-family no-framing baseline from the main run. \\
Framed flat replacement & 94.84 & 34 & 15 & 11 & +4
& A narrow replacement variant can help but is not robust across framing styles. \\
Framed style-balanced replacement & 94.31 & 48 & 16 & 19 & $-3$
& Balanced style prompts already degrade consensus. \\
Framed style+temp replacement & 93.71 & 61 & 19 & 30 & $-11$
& More varied framing further degrades, showing replacement fragility. \\
Guided re-solve replacement & 81.65 & 215 & 6 & 176 & $-170$
& Guided outputs are too noisy to replace consensus directly. \\
Panel comparison replacement & 85.82 & 172 & 16 & 131 & $-115$
& Panel outputs are too noisy to replace consensus directly. \\
Main additive $\Delta$ F+G & 94.84 & 16 & 9 & 5 & +4
& The main fixed policy uses framed and guided evidence as additive side evidence. \\
\bottomrule
\end{tabular}
\end{table}

\subsection{Qwen3-4B GSM8K additive delta policies}

\begin{table}[!htbp]
\centering
\caption{Qwen3-4B GSM8K additive delta source-set policies. Baseline raw consensus is $94.54\%$. The fixed main policy is framed+guided with top-2-mass reliability; panel/order variants are ablations and are not used for the headline matrix.}
\label{tab:app_qwen_delta}
\small
\begin{tabular}{lrrrrr}
\toprule
\textbf{Policy} & \textbf{Acc.} & \textbf{Overrides}
& \textbf{Rec.} & \textbf{Deg.} & \textbf{Net} \\
\midrule
Raw consensus & 94.54 & 0 & 0 & 0 & 0 \\
$\Delta$ framed+guided (main, top-2 mass) & 94.84 & 16 & 9 & 5 & +4 \\
$\Delta$ panel+guided (mass+order) & 95.00 & 20 & 12 & 6 & +6 \\
$\Delta$ panel+guided (mass+order+seed) & 95.00 & 20 & 12 & 6 & +6 \\
$\Delta$ panel+guided (mass) & 94.92 & 23 & 13 & 8 & +5 \\
$\Delta$ raw framed+panel & 94.92 & 25 & 13 & 8 & +5 \\
$\Delta$ all sources (mass+order) & 94.84 & 18 & 10 & 6 & +4 \\
$\Delta$ raw all & 94.69 & 30 & 14 & 12 & +2 \\
\bottomrule
\end{tabular}
\end{table}

\subsection{Answer-visible frame panel ablation}

\begin{table}[!htbp]
\centering
\caption{Answer-visible ablation on Qwen3-4B GSM8K. Showing candidate answers
to the panel improves the best policies in this run.}
\label{tab:app_answer_visible}
\small
\begin{tabular}{lrrrr}
\toprule
\textbf{Policy} & \textbf{Visible acc.} & \textbf{Net}
& \textbf{Hidden acc.} & \textbf{Net} \\
\midrule
mass-order panel+guided & 95.00 & +6 & 94.84 & +4 \\
mass-order-seed panel+guided & 95.00 & +6 & 94.77 & +3 \\
Reference additive & 94.84 & +4 & 94.77 & +3 \\
\bottomrule
\end{tabular}
\end{table}

\subsection{Full delta matrix}

\begin{table}[!htbp]
\centering
\caption{Full \textsc{Arbiter-$\Delta$} matrix for the fixed framed+guided parameter-free policy.}
\label{tab:app_delta_matrix}
\small
\begin{tabular}{llrrrr}
\toprule
\textbf{Model} & \textbf{Dataset} & \textbf{N}
& \textbf{Consensus} & \textbf{\textsc{Arbiter-$\Delta$}} & \textbf{Net} \\
\midrule
Llama-3.1-8B & GSM8K & 1319 & 91.81 & 92.65 & +11 \\
Llama-3.1-8B & MMLU-HS-Math & 270 & 78.52 & 80.00 & +4 \\
Llama-3.1-8B & MATH-500 & 500 & 51.60 & 54.60 & +15 \\
\midrule
Qwen3-4B & GSM8K & 1319 & 94.54 & 94.84 & +4 \\
Qwen3-4B & MMLU-HS-Math & 270 & 94.81 & 95.19 & +1 \\
Qwen3-4B & MATH-500 & 500 & 69.20 & 69.20 & 0 \\
\midrule
Phi-4 & GSM8K & 1319 & 96.13 & 96.44 & +4 \\
Phi-4 & MMLU-HS-Math & 270 & 92.59 & 93.70 & +3 \\
Phi-4 & MATH-500 & 500 & 73.00 & 73.20 & +1 \\
\bottomrule
\end{tabular}
\end{table}

\clearpage
\section{\textsc{Arbiter-Enc} gain ranges}
\label{app:enc_ranges}

\begin{table}[!htbp]
\centering
\caption{Mean extra gain range of \textsc{Arbiter-Enc} over \textsc{Arbiter-$\Delta$}. The endpoints summarize saved fixed/no-tune and calibrated residual variants, not random-seed confidence intervals.}
\label{tab:app_enc_mean}
\small
\begin{tabular}{lr}
\toprule
\textbf{Variant} & \textbf{Mean extra gain over $\Delta$} \\
\midrule
Fixed / no tune & $-0.1$ to $+0.5$ \\
Calibrated & $+0.8$ to $+1.5$ \\
\bottomrule
\end{tabular}
\end{table}

\begin{table}[!htbp]
\centering
\caption{Cell-level extra gain ranges for \textsc{Arbiter-Enc} over \textsc{Arbiter-$\Delta$}. The endpoints summarize saved residual variants, not random-seed confidence intervals.}
\label{tab:app_enc_cell}
\small
\begin{tabular}{llr}
\toprule
\textbf{Model} & \textbf{Dataset} & \textbf{Extra gain over $\Delta$} \\
\midrule
Llama-3.1-8B & GSM8K & $-0.2$ to $+1.0$ \\
Llama-3.1-8B & MMLU-HS-Math & $+1.0$ to $+3.0$ \\
Llama-3.1-8B & MATH-500 & $+0.8$ to $+2.0$ \\
\midrule
Qwen3-4B & GSM8K & $+0.1$ to $+0.6$ \\
Qwen3-4B & MMLU-HS-Math & $\sim0$ \\
Qwen3-4B & MATH-500 & $-0.2$ to $+1.0$ \\
\midrule
Phi-4 & GSM8K & $\sim0$ to $+0.2$ \\
Phi-4 & MMLU-HS-Math & $+0.2$ to $+0.8$ \\
Phi-4 & MATH-500 & $+0.1$ to $+0.6$ \\
\bottomrule
\end{tabular}
\end{table}

\clearpage
\section{Reliability tiers}
\label{app:reliability_tiers}

\begin{table}[!htbp]
\centering
\caption{Reliability tiers for reported results. This prevents clean
zero-external-information results from being confused with calibrated,
hard-slice, or gold-tuned exploratory results.}
\label{tab:app_reliability_tiers}
\footnotesize
\setlength{\tabcolsep}{4pt}
\begin{tabular}{p{4cm}p{9cm}}
\toprule
\textbf{Tier} & \textbf{Results} \\
\midrule
Clean zero-external-information
& Raw consensus; main framed+guided additive policy $94.54\%\rightarrow94.84\%$ on Qwen3-4B GSM8K; full 3$\times$3
\textsc{Arbiter-$\Delta$} matrix with 8 positive-net, 1 neutral-net, 0 negative cells. \\
Clean but not full-coverage accuracy
& Basin encoder Task A abstention: 50\% coverage reaches 98.79\% kept-set
accuracy. \\
Calibrated
& \textsc{Arbiter-Enc-Calibrated}; residual rule calibrated using
labeled validation data. \\
Gold-tuned exploratory
& historical guarded rerank at 95.22\% if the best configuration is selected using
target-test gold accuracy, and the earlier positive self-review audit that did not replicate. These are useful audit records but not strict deployable selectors. \\
Hard-slice positive
& Answer-only two-turn review on the harder slice:
$64.71\%\rightarrow70.59\%$. \\
Negative / diagnostic
& Robustness/resolve, token-1 FFN intervention, branch-token forcing, one-cell
patching, broad self-review, cluster memos, basin-principle judging, broad
top-up merging, GNN/router selection, global framing-first replacement, direct
panel/guided replacement. \\
\bottomrule
\end{tabular}
\end{table}

\clearpage
\section{Worked \textsc{Arbiter-$\Delta$} calculation}
\label{app:worked_delta}

This appendix gives a complete arithmetic example for Eq.~(M2). Consider the wrong-majority case illustrated in Figure~\ref{fig:basin_combined}. The dominant basin has raw support $b_1=18$ and the leading challenger has raw support $b_2=4$. The auxiliary evidence counts are $f_2=13,f_1=8$ for framed-pool solves, $p_2=9,p_1=2$ for panel trials, and $g_2=4,g_1=0$ for guided re-solves. The main policy uses the framed+guided sources $F$ and $G$. With the fixed Laplace smoothing constant $\alpha=1$ and unit reliability for this illustrative calculation,
\begin{flalign}
&& \small \Delta_2
= \log\frac{4+1}{18+1}
+ \log\frac{13+1}{8+1}
+ \log\frac{4+1}{0+1}.
&& \tag{A1}
\end{flalign}
The three terms are $-1.335$, $0.442$, and $1.609$, so $\Delta_2=0.716>0$. The raw prior favors the dominant basin, but the framed and guided same-model evidence streams more than compensate for that prior, so Eq.~(M3) selects the challenger. A full-source ablation would add the panel term $\log((9+1)/(2+1))=1.204$. In actual main-policy runs, the same calculation includes $r_f(q)$ and $r_g(q)$ from Eq.~(M1); source-set ablations with panels also use $\rho_{P,r}$ from Eq.~(M1p). These factors reduce the contribution of fragmented or order-sensitive sources.

\clearpage
\section{Prompt-template summary}
\label{app:prompt_templates}


Full prompt strings are included in the supplementary code package ZIP\@. This appendix summarizes the prompt families so that the evidence budget is understandable without opening the code. The GSM8K raw-consensus script uses four ordinary solve templates: step-by-step solving with a required \texttt{\#\#\#\# <number>} final line, an equation-first variant, a constraints/backwards-reasoning variant, and a solve-then-arithmetic-self-check variant. The script samples across these templates and temperatures, then clusters by the extracted final answer.

The framing-basin runs add three families. First, framed-pool prompts ask the model to identify the target quantity, entities, units, and operation pattern before solving. Second, panel prompts compare two basin interpretations side by side, then ask the same frozen model to solve fresh. Third, guided prompts give one basin interpretation as a hypothesis and ask the same model to verify or reject it by re-solving. For MMLU-HS-Math, prompts require a final A/B/C/D answer; for MATH-500, prompts require a boxed mathematical answer. The supplementary ZIP contains the exact prompt files and scripts used to instantiate these families.

\clearpage
\section{Three-seed replication}
\label{app:three_seed_replication}

Table~\ref{tab:three_seed_replication} reports a related Qwen3-4B GSM8K
\textsc{Arbiter-$\Delta$} source-set diagnostic over three full random seeds. The varying factor
is the sampling/evidence-generation seed; the model, dataset split, parser, and
policy family are held fixed within this diagnostic. The main matrix in
Table~\ref{tab:delta_main} uses the fixed framed+guided policy, so this table is
a seed-variability reference rather than a formal confidence interval for all
main-policy cells.

\begin{table}[!htbp]
\centering
\caption{Three-seed replication table for a related Qwen3-4B GSM8K \textsc{Arbiter-$\Delta$} source-set diagnostic. Error bars in the final row are 1-sigma sample standard deviations over the three full runs.}
\label{tab:three_seed_replication}
\small
\setlength{\tabcolsep}{4pt}
\begin{tabular}{lrrrrrr}
\toprule
\textbf{Seed} & \textbf{Raw cons.} & \textbf{\textsc{Arbiter-$\Delta$}} & \textbf{Gain} & \textbf{Overrides} & \textbf{Rec./Deg.} & \textbf{Net} \\
\midrule
1234 & 94.54 & 95.00 & +0.46 & 20 & 12 / 6 & +6 \\
5678 & 94.24 & 94.92 & +0.68 & 25 & 16 / 7 & +9 \\
0000 & 94.77 & 95.15 & +0.38 & 17 & 11 / 6 & +5 \\
\midrule
Mean $\pm$ std & 94.52 $\pm$ 0.27 & 95.02 $\pm$ 0.12 & +0.51 $\pm$ 0.16 & 20.7 $\pm$ 4.0 & 13.0 $\pm$ 2.6 / 6.3 $\pm$ 0.6 & +6.7 $\pm$ 2.1 \\
\bottomrule
\end{tabular}
\end{table}

\clearpage
\section{Reproducibility, compute, assets, and broader impacts}
\label{app:reproducibility_compute_assets}

\noindent\textbf{Reproducibility.}
All main experiments use frozen public instruction-tuned models, public benchmark test splits, deterministic answer extraction/normalization rules, and explicit evidence budgets. The key reproducibility parameters are the model identifier, dataset split, raw-pool size $K$, prompt family, sampling temperatures, parser/equivalence rule, random seed, and the number of framed, panel, and guided evidence trials. Gold answers are used only after prediction for evaluation. The supplementary code package includes scripts, configs, prompt files, summary artifacts, a README with exact commands, and a package-version lock file.

\noindent\textbf{Seed variability.}
Table~\ref{tab:three_seed_replication} reports the three-seed source-set diagnostic for Qwen3-4B GSM8K. The reported error bars are 1-sigma sample standard deviations over three full runs with different sampling/evidence seeds. We use this as a seed-variability check rather than as a formal hypothesis test for the main matrix.

\noindent\textbf{Compute environment.}
Experiments were run on RunPod H100 GPU instances using the container \texttt{runpod/pytorch:1.0.2-cu1281-torch280-ubuntu2404} (RunPod PyTorch 2.8.0, CUDA 12.8, Ubuntu 24.04). We used vLLM-style batched inference~\citep{kwon2023vllm}, PyTorch~\citep{paszke2019pytorch}, Hugging Face Transformers~\citep{wolf2020transformers}, and Hugging Face Datasets~\citep{lhoest2021datasets}. Package versions were not updated beyond the April 2026 environment used for the reported runs; the exact installed versions should be provided in the supplementary \texttt{requirements\_lock\_pip\_freeze.txt}.

\noindent\textbf{Compute budget.}
A single model--dataset consensus cell records one greedy pass plus $K{=}24$ sampled generations per question; $K$ counts sampled generations, and the run manifest records \texttt{greedy\_generated} and \texttt{greedy\_in\_consensus}. The main \textsc{Arbiter-$\Delta$} policy additionally uses about 24 framed solves and 4 guided re-solves per question; source-set ablations with panels use about 12 additional panel trials. Non-encoder \textsc{Arbiter-$\Delta$} experiments can usually be completed within hours on H100-class hardware; a full GSM8K cell typically takes roughly 8--12+ hours depending on model, decoding budget, and batching. Hidden-state encoder experiments are substantially more expensive because teacher-forced hidden-state extraction dominates runtime; end-to-end encoder runs can require several days. Preliminary negative experiments consumed more total compute than the final selector because they include self-review, top-up, graph-router, trajectory-encoder, and review-ablation branches.

\noindent\textbf{Existing assets and licenses.}
Table~\ref{tab:asset_license_table} lists the main existing assets. We do not redistribute model weights. The supplementary ZIP includes an \texttt{ASSET\_LICENSES.md} file with the same information and any additional package licenses.

\begin{table}[!h]
\centering
\caption{Existing assets, license/terms, and use in the paper.}
\label{tab:asset_license_table}
\footnotesize
\setlength{\tabcolsep}{3pt}
\begin{tabular}{%
>{\raggedright\arraybackslash}p{1.8cm}
>{\raggedright\arraybackslash}p{4.7cm}
>{\raggedright\arraybackslash}p{2.8cm}
>{\raggedright\arraybackslash}p{3.8cm}
}
\toprule
\textbf{Asset} & \textbf{Identifier/version} & \textbf{License or terms} & \textbf{Use} \\
\midrule
GSM8K & \texttt{openai/gsm8k}, main test & MIT & Evaluation benchmark; numeric answer extraction. \\
MMLU-HS-Math & \texttt{cais/mmlu}, high\_school\_mathematics test & MIT & Evaluation benchmark; multiple-choice answer extraction. \\
MATH-500 & \texttt{HuggingFaceH4/MATH-500} & Derived from MATH/PRM800K split; upstream MATH is MIT & Evaluation benchmark; boxed-answer and symbolic-equivalence evaluation. \\
Qwen3-4B-Instruct-2507 & \texttt{Qwen/Qwen3-4B-Instruct-2507} & Apache-2.0 & Frozen base model for generation and evidence collection. \\
Llama-3.1-8B-Instruct & \texttt{meta-llama/Llama-3.1-8B-Instruct} & Llama 3.1 Community License & Frozen base model for generation and evidence collection. \\
Phi-4 & \texttt{microsoft/phi-4} & MIT & Frozen base model for generation and evidence collection. \\
vLLM & April-2026 installed version & Apache-2.0 & Batched inference and sampling. \\
PyTorch & RunPod PyTorch 2.8.0 image & BSD-style & Tensor computation and hidden-state extraction. \\
Transformers / Datasets & April-2026 installed versions & Apache-2.0 & Tokenizers, model loading, and benchmark loading. \\
\bottomrule
\end{tabular}
\end{table}

\noindent\textbf{Broader impacts.}
The work improves selection among a model's own sampled reasoning outputs. Potential positive impacts include more reliable mathematical reasoning and better auditing of why consensus fails. Potential negative impacts follow from improving the reliability of automated reasoning systems: users may over-trust generated answers, or the same selection machinery could improve harmful automated problem solving in domains outside the math benchmarks studied here. We mitigate these risks by reporting recovered/degraded counts, stating that the method is not a correctness verifier, and limiting claims to frozen-model inference on evaluated reasoning benchmarks.

\clearpage
\section{Additional trajectory-graph visualization}

\label{app:hidden_state_graph_fig}

\begin{figure}[!htbp]
\centering
\IfFileExists{fig3_hidden_state_graph_snapshots.pdf}{%
  \includegraphics[width=\linewidth]{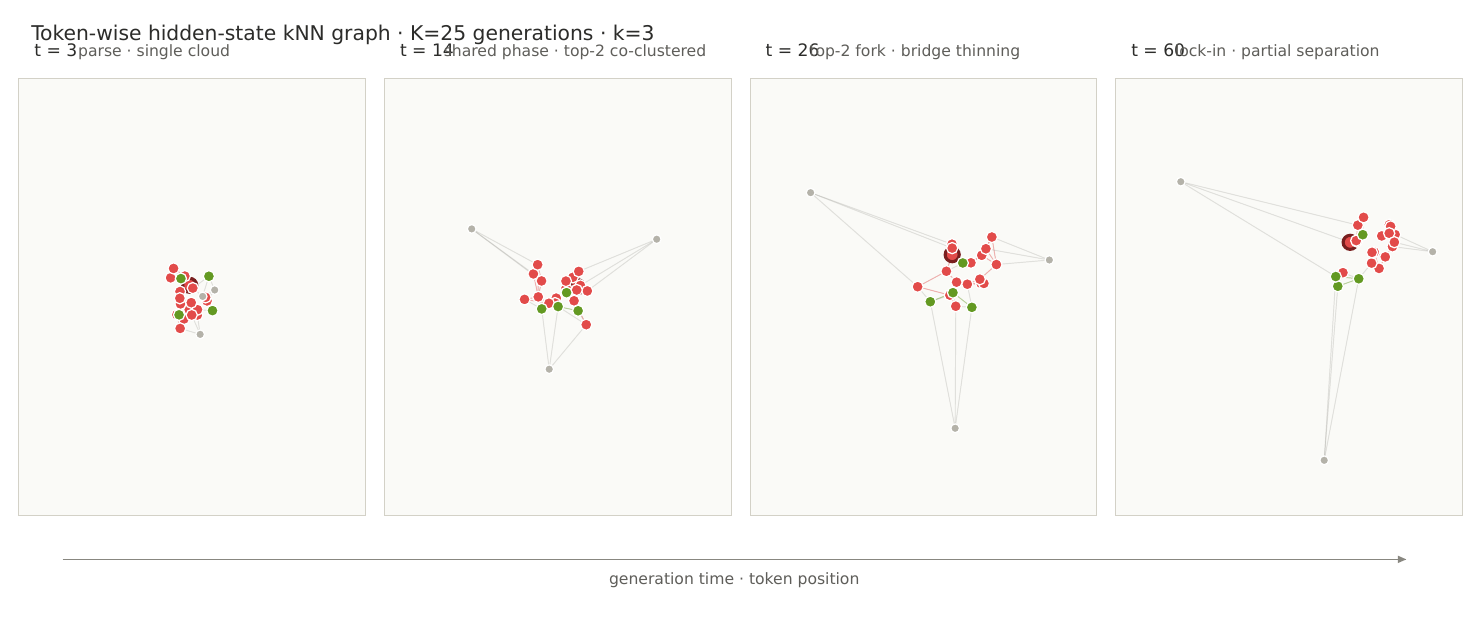}%
}{%
  \fbox{\parbox{0.92\linewidth}{\centering Placeholder for \texttt{fig3\_hidden\_state\_graph\_snapshots.pdf}. Upload the figure file in Overleaf to render this panel.}}%
}
\caption{Token-wise hidden-state kNN graph for the same example. Each panel shows a $k=3$ nearest-neighbor graph over hidden states of 24 sampled generations plus the separately recorded greedy anchor, projected to two dimensions. At early tokens all generations occupy one tight cloud; long-tail basins peel off first; the top two basins become distinguishable but remain partially overlapping. This supports the diagnostic role of trajectory graphs while explaining why hidden-state similarity alone is too graded to serve as a direct correctness selector.}
\label{fig:hidden_state_graph_snapshots}
\end{figure}

\clearpage
\section{Recorded data artifacts}
\label{app:artifact_map}

Table~\ref{tab:app_artifact_map} lists the main artifact families recorded
across the project. We include this table to make the empirical evidence
auditable and reproducible.

\begin{table}[!htbp]
\centering
\caption{Artifact map.}
\label{tab:app_artifact_map}
\footnotesize
\setlength{\tabcolsep}{4pt}
\begin{tabular}{p{3cm}p{10cm}}
\toprule
\textbf{Artifact family} & \textbf{Contents} \\
\midrule
Baseline generation
& \texttt{questions.jsonl}, \texttt{generations.jsonl}, \texttt{clusters.jsonl},
\texttt{pool.jsonl}, \texttt{pool\_summary.json}, \texttt{per\_example.jsonl},
\texttt{summary.json}, \texttt{gate\_sweep.json}. \\
Hidden-state / trajectory
& Zipper checkpoints, feature shards, branch-window tensors, qid-level
trajectory features, resistance curves, branch metadata, merge diagnostics. \\
Text-space steering
& Fragment banks, donor chunk banks, continuation trials, steerability scores,
gate sweeps, step banks. \\
Review sessions
& Session records, follow-up trials, review packets, review outputs, guarded
override decisions, answer-only review trials. \\
Cluster memos
& Representative traces, cluster mutual information, hidden mutual trajectories,
answer-including memos, mix trials, selector sweeps. \\
Basin encoder
& \texttt{basin\_cache.jsonl}, encoder checkpoints, training history,
difficulty scores, abstention curves, review-loop artifacts. \\
Graph router
& Graph shards, router checkpoints, router scores, budget sweeps, graph
diagnostic summaries. \\
Framing / delta
& Baseline pools, framed pools, basin frames, frame-panel trials,
frame-guided trials, policy summaries, delta diagnostics, per-question delta
decisions, matrix summaries. \\
\bottomrule
\end{tabular}
\end{table}

\clearpage
\section{Framing and delta artifact paths}
\label{app:framing_artifacts}

\begin{table}[!htbp]
\centering
\caption{Framing-basin and delta artifacts produced per run or per cell.}
\label{tab:app_framing_artifacts}
\footnotesize
\setlength{\tabcolsep}{4pt}
\begin{tabular}{p{6cm}p{6cm}}
\toprule
\textbf{Artifact} & \textbf{Meaning} \\
\midrule
\texttt{baseline/baseline\_pool.jsonl} & Every baseline generation with extracted answer. \\
\texttt{baseline/baseline\_summary.json} & Baseline consensus metrics. \\
\texttt{framed\_pool/framed\_pool.jsonl} & Framing-first generations. \\
\texttt{framed\_pool/framed\_summary.json} & Framed-pool metrics. \\
\texttt{basin\_frames/basin\_frames.jsonl} & Extracted semantic frames per basin. \\
\texttt{basin\_frame\_summaries\_raw.jsonl} & Raw frame summaries. \\
\texttt{frame\_panel/frame\_panel\_trials.jsonl} & Prompt/output/vote for every panel trial. \\
\texttt{frame\_panel/frame\_panel\_per\_qid.jsonl} & Aggregated panel votes per qid. \\
\texttt{frame\_guided/frame\_guided\_trials.jsonl} & Frame-guided re-solve trials. \\
\texttt{frame\_guided/frame\_guided\_per\_qid.jsonl} & Aggregated guided votes per qid. \\
\texttt{policy\_eval/policy\_summary.json} & Fixed label-free policy ranking. \\
\texttt{policy\_eval/policy\_results.jsonl} & Per-question delta scores and source breakdown. \\
\texttt{delta\_top2/delta\_summary.json} & Delta scorer summary. \\
\texttt{delta\_top2/delta\_qid\_diagnostics.jsonl} & Per-qid delta scores and source breakdown. \\
\texttt{delta\_top2/delta\_policy\_results.jsonl} & Per-question delta policy decisions. \\
\texttt{run\_summary.json} & Full run config and timing, including \texttt{raw\_K}, \texttt{greedy\_generated}, and \texttt{greedy\_in\_consensus}. \\
\bottomrule
\end{tabular}
\end{table}

\clearpage

\end{document}